\pdfoutput=1

\documentclass[11pt]{article}

\usepackage[]{latex/acl}

\usepackage{times}
\usepackage{latexsym}

\usepackage[T1]{fontenc}

\usepackage[utf8]{inputenc}

\usepackage{microtype}

\usepackage{inconsolata}

\usepackage{graphicx} 
\graphicspath{{figs/}}

\usepackage{bm}
\usepackage{amsmath}
\usepackage{amssymb}

\usepackage{bbding}

\usepackage{booktabs}
\usepackage{multicol,multirow}
\usepackage{makecell}

\definecolor{deepred}{rgb}{0.698,0.133,0.133}
\definecolor{blue}{rgb}{0,0,1}

\usepackage{tikz}
\usepackage[edges]{forest}
\usepackage{pgfplots}

\usepackage[edges]{forest}
\usepackage[framemethod=tikz]{mdframed}
\usepackage{subcaption}
\definecolor{lgreen}{rgb}{0.89,0.94,0.85}
\definecolor{lred}{rgb}{0.98, 0.90, 0.84}
\definecolor{lyellow}{rgb}{1.00, 0.95, 0.80}
\definecolor{lblue}{rgb}{0.85, 0.89, 0.95}
\definecolor{hidden-draw}{RGB}{20,68,106}
\definecolor{hidden-pink}{RGB}{255,245,247}

\tikzset{%
    parent/.style =          {align=center,text width=0.7cm, rounded corners=2pt, line width=0.8mm, fill=white!0, draw=white!90},
    child/.style =           {align=center,text width=1.4cm,rounded corners=2pt, fill=blue!10,draw=blue!90,line width=0.3mm},
    T1/.style =           {align=center,text width=1.8cm,rounded corners=3pt, fill=lblue!100, draw=black,line width=0.2mm},   
    T1_end/.style =           {align=left, text width=5cm,rounded corners=5pt, fill=lblue!100,draw=blue!0,line width=0.3mm},
    T2/.style =           {align=center,text width=1.8cm,rounded corners=3pt, fill=lred!100, draw=black,line width=0.2mm},   
    T2_end/.style =           {align=left, text width=5cm,rounded corners=5pt, fill=lred!100,draw=blue!0,line width=0.3mm},
    T3/.style =           {align=center,text width=1.8cm,rounded corners=3pt, fill=lyellow!100, draw=black,line width=0.2mm},   
    T3_end/.style =           {align=left, text width=5cm,rounded corners=5pt, fill=lyellow!100,draw=blue!0,line width=0.3mm},
    T4/.style =           {align=center,text width=1.8cm,rounded corners=3pt, fill=lgreen!100, draw=black,line width=0.2mm},   
    T4_end/.style =           {align=left, text width=5cm,rounded corners=5pt, fill=lgreen!100,draw=blue!0,line width=0.3mm}
}

\title{MM-LLMs: Recent Advances in MultiModal Large Language Models}

\author{Duzhen Zhang\textsuperscript{1}\footnotemark[1]\footnotemark[3] , Yahan Yu\textsuperscript{3}\footnotemark[1] , Jiahua Dong\textsuperscript{4}\footnotemark[2], Chenxing Li\textsuperscript{1} , Dan Su\textsuperscript{1}, \\ \textbf{Chenhui Chu\textsuperscript{3}}\footnotemark[2] \and \textbf{Dong Yu\textsuperscript{2}}\\
  \textsuperscript{1}Tencent AI Lab, China     \textsuperscript{2}Tencent AI Lab, USA \textsuperscript{3}Kyoto University, Japan\\
    \textsuperscript{4}Mohamed bin Zayed University of Artificial Intelligence, United Arab Emirates\\
             \texttt{\{duzhen.zhang972,dongjiahua1995\}}\texttt{@gmail.com}\\\texttt{\{yahan@nlp.ist.,chu@\}}\texttt{i.kyoto-u.ac.jp}, \texttt{\{chenxingli@,dansu@,dyu@global.\}}\texttt{tencent.com}
             }

\begin{document}
\maketitle
\renewcommand{\thefootnote}{\fnsymbol{footnote}}
\footnotetext[1]{Equal contributions.}
\footnotetext[2]{Corresponding authors.}
\footnotetext[3]{This work was done when Duzhen Zhang was interning at Tencent, AI Lab, Beijing, China.}
\renewcommand{\thefootnote}{\arabic{footnote}}

\begin{abstract}
In the past year, MultiModal Large Language Models (MM-LLMs) have undergone substantial advancements, augmenting off-the-shelf LLMs to support MM inputs or outputs via cost-effective training strategies. The resulting models not only preserve the inherent reasoning and decision-making capabilities of LLMs but also empower a diverse range of MM tasks. In this paper, we provide a comprehensive survey aimed at facilitating further research of MM-LLMs. Initially, we outline general design formulations for model architecture and training pipeline. Subsequently, we introduce a taxonomy encompassing $126$ MM-LLMs, each characterized by its specific formulations. Furthermore, we review the performance of selected MM-LLMs on mainstream benchmarks and summarize key training recipes to enhance the potency of MM-LLMs.
Finally, we explore promising directions for MM-LLMs while concurrently maintaining a real-time tracking website\footnote{\url{https://mm-llms.github.io}} for the latest developments in the field. We hope that this survey contributes to the ongoing advancement of the MM-LLMs domain.
\end{abstract}

\section{Introduction}

MultiModal (MM) pre-training research has witnessed significant advancements in recent years, consistently pushing the performance boundaries across a spectrum of downstream tasks~\cite{li2020oscar,akbari2021vatt,fang2021clip2video,yan2021video,li2021align,radford2021learning,li2022blip,zellers2022merlot,zeng2022multi,yang2022vision,wang2022ofa,wang2022image}. However, as the scale of models and datasets continues to expand, traditional MM models incur substantial computational costs, particularly when trained from scratch.
Recognizing that MM research operates at the intersection of various modalities, a logical approach is to capitalize on readily available pre-trained unimodal foundation models, with a special emphasis on powerful Large Language Models (LLMs)~\cite{chatgpt}. This strategy aims to mitigate computational expenses and enhance the efficacy of MM pre-training, leading to the emergence of a novel field: MM-LLMs.

\begin{figure}[t]
\centering
  \includegraphics[width=1.0\linewidth]{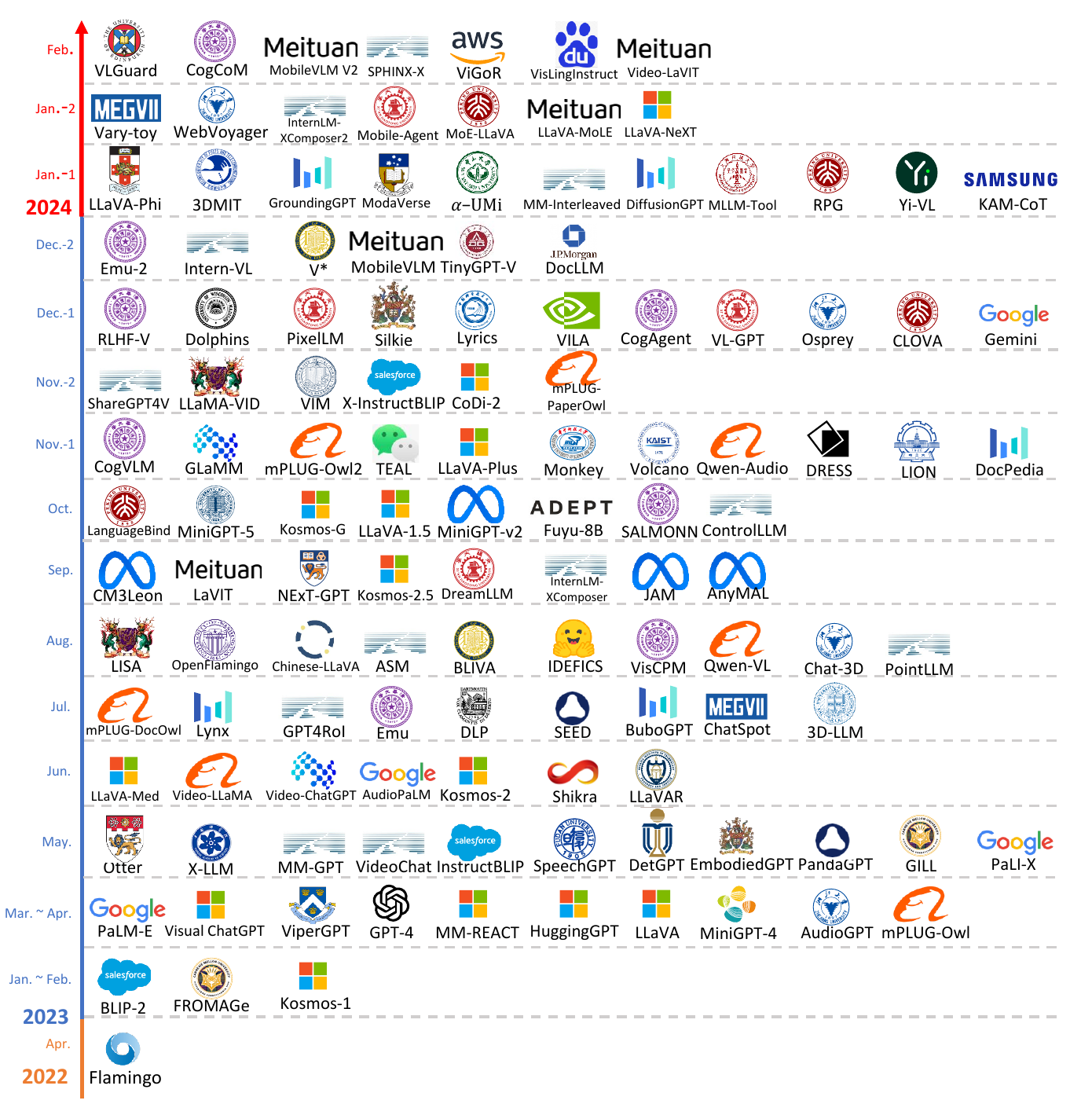} 
    \caption{The timeline of MM-LLMs.}
\label{fig_timeline}
\end{figure}

MM-LLMs harness LLMs as the cognitive powerhouse to empower various MM tasks. LLMs contribute desirable properties like robust language generation, zero-shot transfer capabilities, and In-Context Learning (ICL). Concurrently, foundation models in other modalities provide high-quality representations.
Considering foundation models from different modalities are individually pre-trained, the core challenge facing MM-LLMs is how to effectively connect LLMs with models in other modalities to enable collaborative inference.
The predominant focus within this field has been on refining alignment between modalities and aligning with human intent via a MM Pre-Training (PT) + MM Instruction-Tuning (IT) pipeline.

With the debut of GPT-4(Vision)~\cite{openai2023gpt4} and Gemini~\cite{team2023gemini}, showcasing impressive MM understanding and generation capabilities, a research fervor on MM-LLMs has been sparked. Initial research primarily focuses on MM content comprehension and text generation, encompassing tasks such as image-text understanding, exemplified by projects like BLIP-2~\cite{DBLP:conf/icml/0008LSH23}, LLaVA~\cite{liu2023llava}, MiniGPT-4~\cite{zhu2023minigpt}, and OpenFlamingo~\cite{awadalla2023openflamingo}; video-text understanding, as demonstrated by initiatives such as VideoChat~\cite{li2023videochat}, Video-ChatGPT~\cite{maaz2023video}, and LLaMA-VID~\cite{li2023llama}; and audio-text understanding, as seen in projects like Qwen-Audio~\cite{chu2023qwen}.
Later, the capabilities of MM-LLMs have been expanded to support specific modality outputs. This includes tasks with image-text output, such as GILL~\cite{koh2023generating}, Kosmos-2~\cite{peng2023kosmos}, Emu~\cite{sun2023generative}, and MiniGPT-5~\cite{zheng2023minigpt}; as well as speech/audio-text output, exemplified by projects like SpeechGPT~\cite{DBLP:conf/emnlp/ZhangLZZWZQ23} and AudioPaLM~\cite{rubenstein2023audiopalm}.
Recent research endeavors have focused on mimicking human-like any-to-any modality conversion, shedding light on the path to artificial general intelligence. Some efforts aim to amalgamate LLMs with external tools to reach an approaching any-to-any MM comprehension and generation, such as Visual-ChatGPT~\cite{wu2023visual}, HuggingGPT~\cite{shen2023hugginggpt}, and AudioGPT~\cite{huang2023audiogpt}.
Conversely, to mitigate propagated errors in the cascade system, initiatives like NExT-GPT~\cite{wu2023next}, CoDi-2~\cite{tang2023codi}, and ModaVerse~\cite{wang2024modaverse} have developed end-to-end MM-LLMs of arbitrary modalities. The timeline of MM-LLMs is depicted in Figure~\ref{fig_timeline}.

In this paper, we present a comprehensive survey aimed at facilitating further research of MM-LLMs. To provide readers with a holistic understanding of MM-LLMs, we initially delineate general design formulations from model architecture (Section~\ref{struct}) and training pipeline (Section~\ref{pipe}). We break down the general model architecture into five components: Modality Encoder (Section~\ref{encoder}), Input Projector (Section~\ref{input}), LLM Backbone (Section~\ref{llm}), Output Projector (Section~\ref{output}), and Modality Generator (Section~\ref{decoder}). 
The training pipeline elucidates how to enhance a pre-trained text-only LLM to support MM input or output, primarily consisting of two stages: MM PT (Section~\ref{mmpt}) and MM IT (Section~\ref{mmit}). In that section, we also provide a summary of mainstream datasets for MM PT and MM IT.
Next, we establish a taxonomy encompassing $126$ State-of-the-Art (SOTA) MM-LLMs, each characterized by specific formulations, and summarize their development trends in Section~\ref{model}. 
In Section~\ref{benchmark}, we comprehensively review the performance of major MM-LLMs on mainstream benchmarks and distill key training recipes to enhance the efficacy of MM-LLMs. In Section~\ref{future}, we offer promising directions for MM-LLMs research.
Moreover, we have established a website (\href{https://mm-llms.github.io}{\textcolor{blue}{https://mm-llms.github.io}}) to track the latest progress of MM-LLMs and facilitate crowd-sourcing updates. Finally, we summarize the entire paper in Section~\ref{conclusion} and discuss related surveys on MM-LLMs in Appendix~\ref{ap:rw}. We aspire for our survey to aid researchers in gaining a deeper understanding of this field and to inspire the design of more effective MM-LLMs.

\section{Model Architecture}\label{struct}

\begin{figure*}[t]
\centering
  \includegraphics[width=1.0\linewidth]{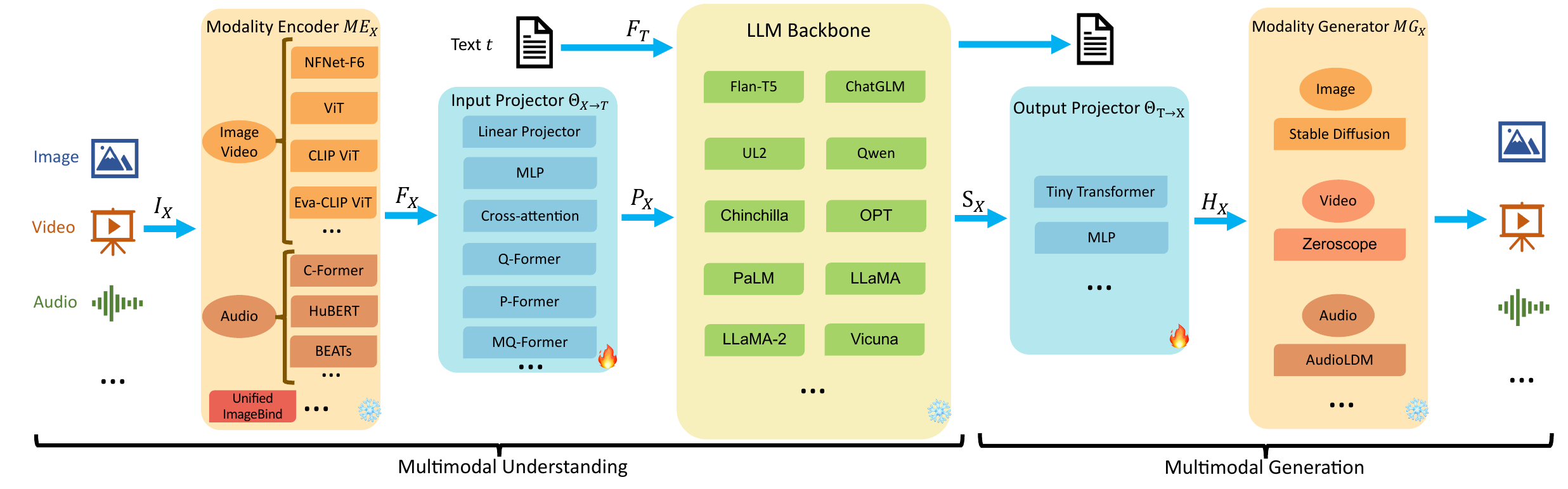} 
    \caption{The general model architecture of MM-LLMs and the implementation choices for each component.}
\label{fig_main}

\end{figure*}

In this section, we provide a detailed overview of the five components comprising the general model architecture, along with the implementation choices for each component, as illustrated in Figure~\ref{fig_main}.
MM-LLMs that emphasize MM understanding only include the first three components.
During training, Modality Encoder, LLM Backbone, and Modality Generator are generally maintained in a frozen state. The primary optimization emphasis is on Input and Output Projectors. 
Given that Projectors are lightweight components, the proportion of trainable parameters in MM-LLMs is notably small compared to the total parameter count (typically around $2$\%). The overall parameter count is contingent on the scale of the core LLM utilized in the MM-LLMs. As a result, MM-LLMs can be efficiently trained to empower various MM tasks.

\subsection{Modality Encoder}\label{encoder}

The Modality Encoder (ME) is tasked with encoding inputs from diverse modalities $I_X$ to obtain corresponding features $\bm{F}_X$, formulated as follows:
\begin{equation}
   \bm{F}_X = \text{ME}_{X}(I_X)\text{.}
\end{equation}

Various pre-trained encoder options $\text{ME}_X$ exist for handling different modalities, where $X$ can be image, video, audio, 3D, etc.
Next, we will offer a concise introduction organized by modality.

\paragraph{Visual Modality}

For images, there are various optional encoders: \textbf{NFNet-F6}~\cite{brock2021high}, \textbf{ViT}~\cite{dosovitskiy2020image}, \textbf{CLIP ViT}~\cite{radford2021learning}, \textbf{Eva-CLIP ViT}~\cite{fang2023eva}, \textbf{BEiT-3}~\cite{wang2023image}, \textbf{OpenCLIP}~\cite{cherti2023reproducible}, \textbf{Grounding-DINO-T}~\cite{zhang2022dino} with Swin-T~\cite{liu2021swin} backbone, \textbf{DINOv2}~\cite{oquab2023dinov2}, \textbf{SAM-HQ}~\cite{kirillov2023segment} with MAE~\cite{he2022masked}, \textbf{RAM++}~\cite{zhang2023recognize} with Swin-B backbone, \textbf{InternViT}~\cite{chen2023internvl}, and \textbf{VCoder}~\cite{jain2023vcoder}. For videos, they can be uniformly sampled to $5$ frames, undergoing the same pre-processing as images.

\paragraph{Audio Modality} is typically encoded by \textbf{C-Former}~\cite{chen2023x}, \textbf{HuBERT}~\cite{hsu2021hubert}, \textbf{BEATs}~\cite{DBLP:conf/icml/ChenW00T0CYW23}, \textbf{Whisper}~\cite{DBLP:conf/icml/RadfordKXBMS23}, and \textbf{CLAP}~\cite{wu2023large}.

\paragraph{3D Point Cloud Modality} is typically encoded by \textbf{ULIP-2}~\cite{Ulip} with a PointBERT~\cite{yu2022point} backbone.

Moreover, to handle numerous heterogeneous modal encoders, some MM-LLMs, particularly any-to-any ones, use \textbf{ImageBind}~\cite{girdhar2023imagebind}, a unified encoder covering six modalities, including image/video, text, audio, heat map, inertial measurement units, and depth.
We provide a brief introduction to some mainstream modality encoders in Appendix~\ref{ap:encoder}.

\subsection{Input Projector}\label{input}

The Input Projector $\mathop{\bm{\Theta}}_{X\to T}$ is tasked with aligning the encoded features of other modalities $\bm{F}_X$ with the text feature space $T$. The aligned features as prompts $\bm{P}_X$ are then fed into the LLM Backbone alongside the textual features $\bm{F}_T$. Given $X$-text dataset $\{I_X, t\}$, the goal is to minimize the $X$-conditioned text generation loss $\mathcal{L}_{\text{txt-gen}}$:
\begin{equation}
    \mathop{\arg\min}_{\bm{\Theta}_{X\to T}}\mathcal{L}_{\text{txt-gen}}(\text{LLM}(\bm{P}_X,\bm{F}_T),t)\text{,}
    \label{optim1}
\end{equation}
where $\bm{P}_X=\mathop{\bm{\Theta}}_{X\to T}(\bm{F}_X)$.

The Input Projector can be achieved directly by a \textbf{Linear Projector} or Multi-Layer Perceptron (\textbf{MLP}), \emph{i.e.}, several linear projectors interleaved with non-linear activation functions. There are also more complex implementations like \textbf{Cross-attention}, \textbf{Q-Former}~\cite{DBLP:conf/icml/0008LSH23}, \textbf{P-Former}~\cite{jian2023bootstrapping}, and \textbf{MQ-Former}~\cite{lu2023lyrics}. \textbf{Cross-attention} (Perceiver Resampler)~\cite{alayrac2022flamingo} uses a set of trainable vectors as queries and the encoded features $\bm{F}_X$ as keys to compress the feature sequence to a fixed length. The compressed representation is then fed directly into the LLM or further used for X-Text cross-attention fusion.
\textbf{Q-Former} extracts relevant features from $\bm{F}_X$ with learnable queries, and the selected features are then used as prompts $\bm{P}_X$. Meanwhile, \textbf{P-Former} generates "reference prompts", imposing an alignment constraint on the prompts produced by Q-Former.
\textbf{MQ-Former} conducts a fine-grained alignment of multi-scale visual and textual signals.
However, both Q-, P-, MQ-Former require an \textbf{additional PT} process for initialization.

\subsection{LLM Backbone}\label{llm}

Taking LLMs~\cite{zhao2023survey,naveed2023comprehensive,mo2024large} as the core agents, MM-LLMs can inherit some notable properties like zero-shot generalization, few-shot ICL, Chain-of-Thought (CoT), and instruction following. The LLM Backbone processes representations from various modalities, engaging in semantic understanding, reasoning, and decision-making regarding the inputs. It produces (1) direct textual outputs $t$, and (2) signal tokens $\bm{S}_X$ from other modalities (if any). These signal tokens act as instructions to guide the generator on whether to produce MM contents and, if affirmative, specifying the content to produce:
\begin{equation}
         t\text{,}\ \bm{S}_X = \text{LLM}(\bm{P}_X, \bm{F}_T)\text{,}
\end{equation}
where the aligned representations of other modalities $\bm{P}_X$ can be considered as soft Prompt-tuning for the LLM.
Moreover, some works have introduced Parameter-Efficient Fine-Tuning (PEFT) methods, such as Prefix-tuning~\cite{li2021prefix}, LoRA~\cite{hu2021lora}, and LayerNorm tuning~\cite{zhao2023tuning}. In these cases, the number of additional trainable parameters is exceptionally minimal, even less than 0.1\% of the total LLM parameter count. We provide an introduction to mainstream PEFT methods in Appendix~\ref{ap:peft}.

The commonly used LLMs in MM-LLMs incude \textbf{Flan-T5}~\cite{chung2022scaling}, 
\textbf{ChatGLM}~\cite{zeng2022glm}, 
\textbf{UL2}~\cite{tay2022ul2}, \textbf{Persimmon}~\cite{persimmon-8b}, \textbf{Qwen}~\cite{bai2023qwen}, \textbf{Chinchilla}~\cite{hoffmann2022training}, \textbf{OPT}~\cite{zhang2022opt}, \textbf{PaLM}~\cite{chowdhery2023palm}, \textbf{LLaMA}~\cite{touvron2023llama}, 
\textbf{LLaMA-2}~\cite{touvron2023llama2}, and \textbf{Vicuna}~\cite{vicuna2023}. We provide a brief introduction to some representative LLMs in Appendix~\ref{ap:model}.

\subsection{Output Projector}\label{output}

The Output Projector $\mathop{\bm{\Theta}}_{T\to X}$ maps the signal token representations $\bm{S}_X$ from the LLM Backbone into features $\bm{H}_X$ understandable to the following Modality Generator $\text{MG}_X$. Given the $X$-text dataset $\{I_X, t\}$, $t$ is first fed into LLM to generate the corresponding $\bm{S}_X$, then mapped into $\bm{H}_X$. To facilitate alignment of the mapped features $\bm{H}_X$, the goal is to minimize the distance between $\bm{H}_X$ and the conditional text representations of $\text{MG}_X$:
\begin{equation}
      \mathop{\arg\min}_{\bm{\Theta}_{T\to X}}\mathcal{L}_{\text{mse}}(\bm{H}_X, \tau_X(t))\text{.}
          \label{optim2}
\end{equation}

The optimization only relies on captioning texts, without utilizing any audio or visual resources $X$, where $\bm{H}_X=\mathop{\bm{\Theta}}_{T\to X}(\bm{S}_X)$ and $\tau_X$ is the textual condition encoder in $\text{MG}_X$.
The Output Projector is implemented by a \textbf{Tiny Transformer} with a learnable decoder feature sequence or \textbf{MLP}.

\subsection{Modality Generator}\label{decoder}

The Modality Generator $\text{MG}_X$ is tasked with producing outputs in distinct modalities. 
Commonly, existing works use off-the-shelf Latent Diffusion Models (LDMs)~\cite{song2021score,bao2022analytic,DBLP:conf/nips/0008BL022}, \emph{i.e.}, \textbf{Stable Diffusion}~\cite{rombach2022high} for image synthesis, \textbf{Zeroscope}~\cite{Cerspense} for video synthesis, and \textbf{AudioLDM-2}~\cite{liuaudioldm2023,liuaudioldm22023} for audio synthesis. The features $\bm{H}_X$ mapped by the Output Projector serve as conditional inputs in the denoising process to generate MM content. 
During training, the ground truth content is first transformed into a latent feature $z_0$ by the pre-trained VAE~\cite{kingma2013auto}. 
Then, noise $\epsilon$ is added to $z_0$ to obtain the noisy latent feature $z_t$. A pre-trained Unet~\cite{ronneberger2015u} $\epsilon_{X}$ is used to compute the conditional LDM loss $\mathcal{L}_{\text{X-gen}}$ as follows:
\begin{equation}
    \mathcal{L}_{\text{X-gen}}:= \mathbb{E}_{\epsilon\sim\mathcal{N}(0,1),t}||\epsilon-\epsilon_{X}(z_t,t,\bm{H}_X)||^2_2\text{,}
        \label{optim3}
\end{equation}
which optimizes parameters $\bm{\Theta}_{X\to T}$ and $\bm{\Theta}_{T\to X}$ by minimizing $\mathcal{L}_{\text{X-gen}}$.

\section{Training Pipeline}\label{pipe}

MM-LLMs' training pipeline can be delineated into two principal stages: MM PT and MM IT.

\tikzstyle{my-box}=[
    rectangle,
    draw=hidden-draw,
    rounded corners,
    text opacity=1,
    minimum height=1.5em,
    minimum width=5em,
    inner sep=2pt,
    align=center,
    fill opacity=.5,
    line width=0.8pt,
]
\tikzstyle{leaf}=[my-box, minimum height=1.5em,
    fill=hidden-pink!80, text=black, align=left,font=\normalsize,
    inner xsep=2pt,
    inner ysep=4pt,
    line width=0.8pt,
]
\begin{figure*}[t!]
    \centering
    \resizebox{1.0\textwidth}{!}{
        \begin{forest}
            forked edges,
            for tree={
                grow'=0,
                draw,
                reversed=true,
                anchor=base west,
                parent anchor=east,
                child anchor=west,
                base=left,
                font=\large,
                rectangle,
                rounded corners,
                align=left,
                minimum width=4em,
                edge+={darkgray, line width=1pt},
                s sep=3pt,
                inner xsep=2pt,
                inner ysep=3pt,
                line width=0.8pt,
                ver/.style={rotate=90, child anchor=north, parent anchor=south, anchor=center},
            },
            where level=1{text width=8em,font=\normalsize,}{},
            where level=2{text width=5.15em,font=\normalsize,}{},
            where level=3{text width=8em,font=\normalsize,}{},
            where level=4{text width=5em,font=\normalsize,}{},
			[
			    MM-LLMs, ver
			    [
			      Functional Division
			        [
			        MM Unders.
			            [
                        \textcolor{deepred}{\textbf{I+T$\to$T:}} BLIP-2~\cite{DBLP:conf/icml/0008LSH23}{, }Kosmos-1~\cite{huang2023language}{, }PaLM-E~\cite{driess2023palm}{, }
                        ViperGPT\\\cite{suris2023vipergpt}{, }LLaVA~\cite{liu2023llava}{, }MiniGPT-4~\cite{zhu2023minigpt}{, }mPLUG-Owl~\cite{ye2023mplug}{, }\\Otter~\cite{li2023otter}{, }MultiModal-GPT~\cite{gong2023multimodal}{, }
                        PandaGPT~\cite{su2023pandagpt}{, }PaLI-X(\citeauthor{chen2023pali})\\LLaVA-Med~\cite{li2023llava}{, }LLaVAR~\cite{zhang2023llavar}{, }mPLUG-DocOwl({$\textbf{I}_\textbf{D}$})~\cite{ye2023mplugdoc}{, }DLP\\\cite{jian2023bootstrapping}{, }ChatSpot~\cite{zhao2023chatspot}{, }OpenFlamingo~\cite{awadalla2023openflamingo}{, }Chinese-LLaVA\\\cite{chinese-llava}{, }ASM~\cite{wang2023all}{, }BLIVA~\cite{hu2023bliva}{, }IDEFICS~\cite{IDEFICS}{, }Qwen-VL\\\cite{Qwen}{, }Kosmos-2.5~\cite{lv2023kosmos}{, }InternLM-XComposer~\cite{zhang2023internlm}{, }JAM\\(\citeauthor{aiello2023jointly}){, }LLaVA-1.5~\cite{liu2023improved}{, }MiniGPT-v2~\cite{chen2023minigpt}{, }Fuyu-8B~\cite{fuyu-8b}{, }\\CogVLM\cite{wang2023cogvlm}{, }mPLUG-Owl2~\cite{ye2023mplug2}{, }Monkey~\cite{li2023monkey}{, }Volcano\\\cite{lee2023volcano}{, }DRESS~\cite{chen2023dress}{, }LION~\cite{chen2023lion}{, }DocPedia({$\textbf{I}_\textbf{D}$})~\cite{feng2023docpedia}{, }\\ShareGPT4V\cite{chen2023sharegpt4v}{, }VIM~\cite{lu2023vim}{, }mPLUG-PaperOwl({$\textbf{I}_\textbf{D}$})\cite{hu2023mplug}{, }RLHF-V\\\cite{yu2023rlhf}{, }Silkie~\cite{li2023silkie}{, }Lyrics~
                        \cite{lu2023lyrics}{, }VILA~\cite{lin2023vila}{, }CogAgent\\\cite{hong2023cogagent}{, }Osprey~\cite{yuan2023osprey}{, }V*~\cite{wu2023v}{, }MobileVLM~\cite{chu2023mobilevlm}{, }\\TinyGPT-V~(\citeauthor{yuan2023tinygpt}){, }DocLLM({$\textbf{I}_\textbf{D}$})~\cite{wang2023docllm}{, }LLaVA-$\phi$~\cite{zhu2024llava}{, }Yi-VL\cite{yi-vl}\\KAM-CoT(\citeauthor{mondal2024kam}){, }InternLM-XComposer2~\cite{dong2024internlm}{, }MoE-LLaVA~\cite{lin2024moe}{, }\\LLaVA-MoLE~\cite{chen2024llava}{,}
                        LLaVA-NeXT~\cite{liu2024llavanext}{, }VLGuard~\cite{zong2024safety}{, }\\MobileVLM V2~\cite{chu2024mobilevlm2}{, }ViGoR\cite{yan2024vigor}{, }VisLingInstruct~\cite{zhu2024vislinginstruct}\\
                          \textcolor{deepred}{\textbf{V+T$\to$T:}} VideoChat~\cite{li2023videochat}{, }Video-ChatGPT~\cite{maaz2023video}{, }Dolphins~\cite{ma2023dolphins}\\
                          \textcolor{deepred}{\textbf{A+T$\to$T:}} SALMONN~\cite{tang2023salmonn}{, }Qwen-Audio~\cite{chu2023qwen}\\
                        \textcolor{deepred}{\textbf{3D+T$\to$T:}} 3D-LLM~\cite{hong20233d}{, }Chat-3D~\cite{wang2023chat3d1}{, }PointLLM~\cite{xu2023pointllm}{, }\\3DMIT~\cite{li20243dmit}\\
                          \textcolor{deepred}{\textbf{Many$\to$T:}} Flamingo~\cite{alayrac2022flamingo}{, }MM-REACT~\cite{yang2023mm}{, }X-LLM~\cite{chen2023x}{, }\\
                        InstructBLIP~\cite{DBLP:journals/corr/abs-2305-06500}{, }EmbodiedGPT~\cite{mu2023embodiedgpt}{, }Video-LLaMA~\cite{DBLP:conf/emnlp/ZhangLB23}{, }Lynx\\\cite{zeng2023matters}{, }AnyMAL\cite{moon2023anymal}{, }LanguageBind~\cite{zhu2023languagebind}{, }LLaMA-VID\\\cite{li2023llama}{, }X-InstructBLIP~\cite{panagopoulou2023x}{, }InternVL~\cite{chen2023internvl}
                            , leaf, text width=46em
			            ]
			        ]
			        [
			      MM Genera.
			            [  \textcolor{deepred}{\textbf{I+T$\to$I+T:}} FROMAGe({$\textbf{I}_\textbf{R}$})~\cite{koh2023grounding}{, }Visual ChatGPT~\cite{wu2023visual}{, }DetGPT({$\textbf{I}_\textbf{B}$})\cite{pi2023detgpt}\\GILL\cite{koh2023generating}{, }Kosmos-2({$\textbf{I}_\textbf{B}$})~\cite{peng2023kosmos}{, }Shikra({$\textbf{I}_\textbf{B}$})~\cite{chen2023shikra}{, }GPT4RoI({$\textbf{I}_\textbf{B}$})\\\cite{zhang2023gpt4roi}{, }SEED~\cite{ge2023planting}{, }LISA({$\textbf{I}_\textbf{M}$})~\cite{lai2023lisa}{, }VisCPM\cite{hu2023large}{, }\\CM3Leon\cite{yu2023scaling}{, }LaVIT~\cite{jin2023unified}{, }DreamLLM~\cite{dong2023dreamllm}{, }MiniGPT-5\\\cite{zheng2023minigpt}{, }Kosmos-G~\cite{pan2023kosmos}{, }GLaMM({$\textbf{I}_\textbf{M}$})~\cite{rasheed2023glamm}{, }LLaVA-Plus(\textbf{+}{$\textbf{I}_\textbf{B}$\&$\textbf{I}_\textbf{M}$})\\\cite{liu2023llavaplus}{, }
               PixelLM({$\textbf{I}_\textbf{M}$})~\cite{ren2023pixellm}{, }VL-GPT~\cite{zhu2023vl}{, }CLOVA(\textbf{+}{$\textbf{I}_\textbf{B}$\&$\textbf{I}_\textbf{M}$})\\\cite{gao2023clova}{, }Emu-2~\cite{sun2023generative2}{, }MM-Interleaved~\cite{tian2024mm}{, }DiffusionGPT\\\cite{qin2024diffusiongpt}{, }RPG\cite{yang2024mastering}{,}Vary-toy({$\textbf{I}_\textbf{B}$})~\cite{wei2024small}{, }CogCoM({$\textbf{I}_\textbf{B}$})~\cite{qi2024cogcom}{,}\\SPHINX-X({$\textbf{I}_\textbf{B}$})~\cite{gao2024sphinx}\\
                \textcolor{deepred}{\textbf{V+T$\to$V+T:}} Video-LaVIT~\cite{jin2024video}\\
                          \textcolor{deepred}{\textbf{A/S+T$\to$A/S+T:}} SpeechGPT~\cite{DBLP:conf/emnlp/ZhangLZZWZQ23}{, }AudioPaLM~\cite{rubenstein2023audiopalm}\\ 
                          \textcolor{deepred}{\textbf{Many$\to$I+T:}} Emu~\cite{sun2023generative}{, }BuboGPT({$\textbf{I}_\textbf{M}$})~\cite{zhao2023bubogpt}{, }GroundingGPT({$\textbf{I}_\textbf{B}$})~\cite{li2024lego}\\
                          \textcolor{deepred}{\textbf{Many$\to$Many:}} GPT-4~\cite{openai2023gpt4}{, }HuggingGPT~\cite{shen2023hugginggpt}{, }AudioGPT~\cite{huang2023audiogpt}\\NExT-GPT~\cite{wu2023next}{, }ControlLLM~\cite{liu2023controlllm}{, }TEAL~\cite{yang2023teal}{, }CoDi-2(\citeauthor{tang2023codi2})\\Gemini~\cite{team2023gemini}{, }ModaVerse~\cite{wang2024modaverse}{, }MLLM-Tool\cite{wang2024tool}\\
                            , leaf, text width=46em
			            ]
			        ]
			    ]
			    [
		           Design Division
			        [
			         Tool-using
			            [Visual ChatGPT\cite{wu2023visual}{, }  ViperGPT\cite{suris2023vipergpt}{, }MM-REACT\cite{yang2023mm}{, }\\
               HuggingGPT~\cite{shen2023hugginggpt}{,}AudioGPT~\cite{huang2023audiogpt}{, }ControlLLM~\cite{liu2023controlllm}{, }LLaVA-Plus\\(\citeauthor{liu2023llavaplus}){, }CogAgent~\cite{hong2023cogagent}{,}CLOVA~\cite{gao2023clova}{, }$\alpha$-UMi~\cite{shen2024small}{,}MLLM-Tool\\(\citeauthor{wang2024tool}){, }WebVoyager~\cite{he2024webvoyager}{, }Mobile-Agent~\cite{wang2024mobile}
			            , leaf, text width=46em
			            ]
			        ]
			        [
			      End-to-end
			            [The remaining models are essentially all end-to-end trainable models.
			              , leaf, text width=46em
			            ]
			        ]
			    ]
			]
            \end{forest}
    }
    \caption{Taxonomy for MM-LLMs. I: Image, V: Video, A/S: Audio/Speech, and T: Text. {$\textbf{I}_\textbf{D}$}: Document understanding, {$\textbf{I}_\textbf{B}$}: Output bounding box, {$\textbf{I}_\textbf{M}$}: Output segmentation mask, and {$\textbf{I}_\textbf{R}$}: Output retrieved images.}
    \label{fig:taxonomy}
\end{figure*}

\subsection{MM PT}\label{mmpt}
During the PT stage, typically leveraging the X-Text datasets, Input and Output Projectors are trained to achieve alignment among various modalities by optimizing predefined objectives.
For MM understanding models, optimization focuses solely on Equation~(\ref{optim1}), while for MM generation models, optimization involves Equations~(\ref{optim1}), (\ref{optim2}), and (\ref{optim3}). In the latter case, Equation~(\ref{optim1}) also includes the ground-truth signal token sequence.

The X-Text datasets include Image-Text, Video-Text, and Audio-Text, with Image-Text having two types: Image-Text pairs (\emph{e.g.}, \textbf{<img1>} <txt1>) and interleaved Image-Text corpus (\emph{e.g.}, <txt1>\textbf{<img1>}<txt2><txt3>\textbf{<img2>}<txt4>). Details of X-Text datasets are shown in Table~\ref{PT_corpus}.

\begin{table*}[t]
  \centering
\resizebox{1.0\linewidth}{!}{
    \begin{tabular}{lcccccccc}
    \toprule
    \textbf{Model} & \textbf{I$\to$O} & \textbf{Modality Encoder} & \textbf{Input Projector} & \textbf{LLM Backbone} & \textbf{Output Projector} & \textbf{Modality Generator} & \textbf{\#.PT}& \textbf{\#.IT} \\
    \midrule
    Flamingo & I+V+T$\to$T& I/V: NFNet-F6& Cross-attention&Chinchilla-1.4B/7B/70B &--&--&--&-- \\
    BLIP-2 & I+T$\to$T& I: CLIP/Eva-CLIP ViT@224& Q-Former w/ Linear Projector&Flan-T5/OPT &--&--&129M&-- \\
    LLaVA & I+T$\to$T& I: CLIP ViT-L/14& Linear Projector&Vicuna-7B/13B &--&--&--&-- \\
    MiniGPT-4 & I+T$\to$T& I: Eva-CLIP ViT-G/14 & Q-Former w/ Linear Projector&Vicuna-13B &--&--&--&-- \\
    mPLUG-Owl & I+T$\to$T& I: CLIP ViT-L/14 & Cross-attention&LLaMA-7B&--&--&--&-- \\
    Otter & I+T$\to$T& I: CLIP ViT-L/14 & Cross-attention & LLaMA-7B & -- & -- & -- & --\\
    X-LLM & I+V+A+T$\to$T& I/V: ViT-G; A: C-Former& Q-Former w/ Linear Projector&ChatGLM-6B &--&--&--&-- \\
   VideoChat & V+T$\to$T& I: ViT-G & Q-Former w/ Linear Projector&Vicuna &--&--&--&-- \\
    InstructBLIP & I+V+T$\to$T& I/V: ViT-G/14@224 & Q-Former w/ Linear Projector&Flan-T5/Vicuna &--&--&129M&1.2M \\
    PandaGPT & I+T$\to$T& I: ImageBind & Linear Projector&Vicuna-13B &--&--&--&-- \\
    GILL & I+T$\to$I+T& I: CLIP ViT-L & Linear Projector & OPT-6.7B & Tiny Transformer & I: Stable Diffusion-1.5 & -- & --\\
    PaLI-X & I+T$\to$T& I: ViT &  Linear Projector&UL2-32B&--&--&--&-- \\
    Video-LLaMA & I+V+A+T$\to$T& \makecell{I/V: Eva-CLIP ViT-G/14;\\ A: ImageBind} & Q-Former w/ Linear Projector&Vicuna/LLaMA&--&--&--&-- \\
    Video-ChatGPT & V+T$\to$T& I: CLIP ViT-L/14& Linear Projector&Vicuna-v1.1&--&--&--&-- \\
    Shikra & I+T$\to$T+$\text{I}_\text{B}$& I: CLIP ViT-L/14@224 & Linear Projector&Vicuna-7B/13B&--&--&600K&5.5M \\
    LLaVAR & I+T$\to$T& \makecell{I: CLIP ViT-L/14@224 \\\& CLIP ViT-L/14@336} & Linear Projector & Vicuna-13B & -- & -- &--&--\\
    mPLUG-DocOwl & $\text{I}_\text{D}$+T$\to$T& I: CLIP ViT-L/14 & Cross-attention & LLaMA-7B & -- & -- &--&--\\
    Lynx & I+V+T$\to$T& I/V: Eva-CLIP ViT-1B & Cross-attention & Vicuna & -- & --&--&-- \\
    Emu & I+V+T$\to$I+T& I/V: Eva-CLIP-1B & Cross-attention & LLaMA-13B & MLP & I: Stable Diffusion-1.5 &--&--\\
   DLP & I+T$\to$T& I: CLIP/Eva-CLIP ViT & Q-\&P-Former w/ Linear Projector&OPT/Flan-T5 &--&--&--&-- \\
    BuboGPT & I+A+T$\to$T+$\text{I}_\text{M}$& \makecell{I: CLIP/Eva-CLIP ViT; \\A: ImageBind} & Q-Former w/ Linear Projector&Vicuna &--&--&--&-- \\
ChatSpot & I+T$\to$T& I: CLIP ViT-L/14 & Linear Projector&Vicuna-7B/LLaMA &--&--&--&-- \\
IDEFICS & I+T$\to$T& I: OpenCLIP & Cross-attention & LLaMA & -- & --& -- & --\\
   Qwen-VL-(Chat) & I+T$\to$T& \makecell{I: ViT@448 initialized\\from OpenClip's ViT-bigG} & Cross-attention&Qwen-7B&--&--&1.4B$^{\dagger}$&50M$^{\dagger}$ \\
   LaVIT & I+T$\to$I+T& I: ViT & Cross-attention & LLaMA-7B & -- & I: Stable Diffusion& -- & --\\
   NExT-GPT & I+V+A+T$\to$I+V+A+T& I/V/A: ImageBind & Linear Projector&Vicuna-7B&Tiny Transformer&\makecell{I: Stable Diffusion; V: Zeroscope;\\A: AudioLDM}&--&-- \\
   DreamLLM & I+T$\to$I+T& I: CLIP ViT-L & Linear Projector & Vicuna & MLP & I: Stable Diffusion& -- & --\\
   AnyMAL & I+V+A+T$\to$T& \makecell{I: CLIP ViT/L\&ViT-G\&DinoV2;\\V: Intervideo; A: CLAP} & \makecell{I/V: Cross-attention;\\A: Linear Projector} & LLaMA-2 & --&--& -- & --\\
  MiniGPT-5 & I+T$\to$I+T& I: Eva-CLIP ViT-G/14 & Q-Former w/ Linear Projector&Vicuna-7B &Tiny Transformer&I: StableDiffusion-2&--&-- \\
    LLaVA-1.5 & I+T$\to$T& I: CLIP ViT-L@336 &MLP&Vicuna-v1.5-7B/13B &--&--&0.6M&0.7M \\
MiniGPT-v2 & I+T$\to$T& I: Eva-CLIP ViT@448 & Linear Projector&LLaMA-2-Chat-7B &--&--&--&-- \\
   CogVLM & I+T$\to$T& I: Eva-2-CLIP ViT & MLP&Vicuna-v1.5-7B &--&--&--&-- \\
   Qwen-Audio & A+T$\to$T& A: Whisper-L-v2 & Linear Projector & Qwen-7B & --&--& -- & --\\
   DRESS & I+T$\to$T& I:Eva-CLIP ViT-G/14 & Linear Projector&Vicuna-v1.5-13B &--&--&--&-- \\
X-InstructBLIP & I+V+A+3D+T$\to$T&\makecell{I/V: Eva-CLIP ViT-G/14;\\A: BEATs; 3D: ULIP-2} & Q-Former w/ Linear Projector&Vicuna-v1.1-7B/13B &--&--&--&-- \\
   CoDi-2 & I+V+A+T$\to$I+V+A+T& I/V/A: ImageBind & MLP&LLaMA-2-Chat-7B &MLP&\makecell{I: Stable Diffusion-2.1;\\V: Zeroscope-v2; A: AudioLDM-2}&--&-- \\
   RLHF-V & I+T$\to$T& I: BEiT-3 & Linear Projector & Vicuna-v1-13B & --&--& -- & --\\
   Silkie & I+T$\to$T&\makecell{I: ViT initialized from\\OpenCLIP's ViT-bigG} & Cross-attention & Qwen-7B & --&--& -- & --\\
   Lyrics & I+T$\to$T& \makecell{I: CLIP ViT-L/14\&Grounding-DINO-T\\\&SAM-HQ\&ViT-H\&RAM++} & MQ-Former w/ Linear Projection & Vicuna-13B &--&--& -- & --\\
VILA & I+T$\to$T& I: ViT@336 & Linear Projector&LLaMA-2-7B/13B &--&--&50M&1M \\
IntrenVL & I+V+T$\to$T& I/V: InternViT-6B; T: LLaMA-7B & Cross-attention w/ MLP & QLLaMA-8B \& Vicuna-13B &--&-- & -- & --\\
ModaVerse & I+V+A+T$\to$I+V+A+T& ImageBind & Linear Projector & LLaMA-2 & MLP & \makecell{I: Stable Diffusion; V: Videofusion;\\A: AudioLDM}& -- & -- \\
MM-Interleaved & I+T$\to$I+T& I: CLIP ViT-L/14 & Cross-attention & Vicuna-13B & Tiny Transformer & I: Stable Diffusion-2.1  & -- & -- \\
    \bottomrule
  \end{tabular}
  }
  \caption{The summary of $43$ mainstream MM-LLMs. I$\to$O: Input to Output Modalities, I: Image, V: Video, A: Audio, 3D: Point Cloud, and T: Text. In Modality Encoder, ``-L'' represents Large, ``-G'' represents Giant, ``/$14$'' indicates a patch size of $14$, and ``@$224$'' signifies an image resolution of $224\times224$. \textbf{\#.PT} and \textbf{\#.IT} represent the scale of dataset during MM PT and MM IT, respectively. $^{\dagger}$ includes in-house data that is not publicly accessible.}
  \label{tab:models}
\end{table*}

\subsection{MM IT}\label{mmit}

MM IT is a method that entails fine-tuning of pre-trained MM-LLMs using instruction-formatted datasets~\cite{wei2021finetuned}. Through this process, MM-LLMs can generalize to unseen tasks by adhering to new instructions, thereby enhancing zero-shot performance. This straightforward yet impactful concept has catalyzed subsequent success in the field of NLP, exemplified by works such as InstructGPT~\cite{ouyang2022training}, OPT-IML~\cite{iyer2022opt}, and InstructBLIP~\cite{DBLP:journals/corr/abs-2305-06500}.

MM IT comprises Supervised Fine-Tuning (SFT) and Reinforcement Learning from Human Feedback (RLHF), aiming to align with human intents and enhance the interaction capabilities of MM-LLMs. SFT converts part of the PT stage data into an instruction-aware format. Using visual Question-Answer (QA) as an example, various templates may be employed like \textbf{(1) "<Image>\{Question\}"} A short answer to the question is; \textbf{(2) "<Image>"} Examine the image and respond to the following question with a brief answer: \textbf{"\{Question\}.} \textbf{Answer:"}; and so on. Next, it fine-tunes pre-trained MM-LLMs using the same optimization objectives. SFT datasets can be structured as either single-turn QA or multi-turn dialogues.

After SFT, RLHF involves further fine-tuning of the model, relying on feedback regarding the MM-LLMs' responses (\emph{e.g.}, Natural Language Feedback (NLF) labeled manually or automatically)~\cite{sun2023aligning}. This process employs a reinforcement learning algorithm to effectively integrate the non-differentiable NLF. The model is trained to generate corresponding responses conditioned on the NLF~\cite{chen2023dress,akyurek2023rl4f}. The statistics for SFT and RLHF datasets are presented in Table~\ref{IT_corpus} of Appendix~\ref{ap:corpus}.

The datasets used by existing MM-LLMs in the MM PT and MM IT stages are diverse, but they are all \textbf{subsets} of the datasets in Tables~\ref{PT_corpus} and~\ref{IT_corpus}.

\section{SOTA MM-LLMs}\label{model}

As shown in Figure~\ref{fig:taxonomy}, we classify the $126$ SOTA MM-LLMs from both functional and design perspectives.
In the design division, ``Tool-using'' denotes treating the LLM as black box and providing access to certain MM expert systems to perform specific MM tasks via reasoning, while ``End-to-End'' signifies that the entire model is trained jointly in an end-to-end manner. Based on the previously defined design formulations, we also conduct a comprehensive comparison of the architectures and training dataset scales for $43$ of these SOTA MM-LLMs, as illustrated in Table~\ref{tab:models}. 
Next, we will summarize their developmental trends and briefly introduce the core contributions of some representative models in Appendix~\ref{ap:sota}.

\begin{table*}[t]
\centering
\resizebox{1.0\linewidth}{!}{
\begin{tabular}{l c | ccc ccccc ccccccc c ccc}
\toprule
\textbf{Model} & \textbf{LLM Backbone}  & \textbf{OKVQA} & \makecell{\textbf{IconVQA}}& \textbf{VQA$^\text{v2}$} & \textbf{GQA} & \makecell{\textbf{VizWiz}} & \textbf{SQA$^\text{I}$} & \makecell{\textbf{VQA$^\text{T}$}} & \textbf{POPE} & \textbf{MME$^{\text{P}}$}&\textbf{MME$^{\text{C}}$} & \textbf{MMB} & \textbf{MMB$^\text{CN}$} & \textbf{SEED$^{\text{I}}$} & \textbf{LLaVA$^\text{W}$} & \textbf{MM-Vet} & \textbf{QBench}& \makecell{\textbf{HM}} & \makecell{\textbf{VSR}}\\
\midrule
Flamingo & Chinchilla-7B & 44.7 & -- & -- & -- & 28.8 & -- & -- & -- & -- & -- & -- & -- & -- & -- & -- & -- & 57.0 & 31.8 \\
BLIP-2 & Flan-T5$_{\text{XXL}} (13B)$ & 45.9 & 40.6 & 65.0 & 44.7 & 19.6 & 61.0 & 42.5 & 85.3 & 1293.8 & 290.0 &-- & -- & 46.4 & 38.1 & 22.4 & -- & 53.7 & 50.9\\
LLaVA & Vicuna-13B & 54.4 & 43.0 & -- & 41.3 & -- & -- & 38.9 & -- & -- & -- & -- & -- & -- & -- & -- & -- & -- & 51.2\\
MiniGPT-4 & Vicuna-13B & 37.5 & 37.6 & -- & 30.8 & -- & -- & 19.4 & -- & -- & -- & -- & -- & -- & -- & -- & -- & -- & 41.6\\
InstructBLIP & Vicuna-7B & -- & -- & -- & 49.2 & 34.5 & 60.5 & 50.1 & -- & --& -- & 36.0 & 23.7 & 53.4 & 60.9 & 26.2 & 56.7& -- & -- \\
InstructBLIP & Vicuna-13B & -- & 44.8 & -- & 49.5 & 33.4 & 63.1 & 50.7 & 78.9 & 1212.8& 291.8 & -- & -- & -- & 58.2 & 25.6 & -- & 57.5 & 52.1 \\
Shikra & Vicuna-13B & 47.2 & -- & 77.4$^*$ & -- & -- & -- & -- & -- & -- & -- &58.8 & -- & -- & -- & -- &54.7& -- & -- \\
IDEFICS-9B & LLaMA-7B & -- & -- & 50.9 & 38.4 & 35.5 & -- & 25.9 & -- & -- & -- &48.2 & 25.2 & -- & -- & -- & -- & -- & -- \\
IDEFICS-80B& LLaMA-65B & -- & -- & 60.0 & 45.2 & 36.0 & -- & 30.9 & -- & -- & -- &54.5 & 38.1 & -- & -- & -- & --& -- & -- \\
Qwen-VL & Qwen-7B & -- & -- & 78.8$^*$ & 59.3$^*$ & 35.2 & 67.1 & 63.8 & -- & --& -- & 38.2 & 7.4 & 56.3 & -- & -- &59.4& -- & -- \\
Qwen-VL-Chat & Qwen-7B& -- & -- & 78.2$^*$ & 57.5$^*$ & 38.9 & 68.2 & 61.5 & -- & 1487.5 & \textcolor{blue}{\textbf{360.7}} & 60.6 & 56.7 & 58.2 & -- & -- & -- & -- & -- \\
LLaVA-1.5 & Vicuna-1.5-7B & -- & -- & 78.5$^*$ & 62.0$^*$ & 50.0 & 66.8 & 58.2 & \textcolor{deepred}{\textbf{85.9}} & 1510.7 & 316.1$^{\ddagger}$ & 64.3 & 58.3 & 58.6 & 63.4 & 30.5 & 58.7 & -- & -- \\ 
\quad +ShareGPT4V & Vicuna-1.5-7B   & -- & -- & \textcolor{blue}{\textbf{80.6}} & -- & 57.2 & 68.4 & -- & -- & \textcolor{blue}{\textbf{1567.4}} & \textcolor{deepred}{\textbf{376.4}} & 68.8 & 62.2 & \textcolor{deepred}{\textbf{69.7}} & 72.6 & 37.6 & \textcolor{deepred}{\textbf{63.4}} & -- & -- \\
LLaVA-1.5 & Vicuna-1.5-13B & -- & -- & {80.0}$^*$ & {\textcolor{deepred}{\textbf{63.3}}}$^*$ & 53.6 & {71.6} & 61.3 & \textcolor{deepred}{\textbf{85.9}} & 1531.3 & 295.4$^{\ddagger}$ & 67.7 & {63.6} & {61.6} & {70.7} & {35.4} & \textcolor{blue}{\textbf{62.1}} & -- & -- \\ 
MiniGPT-v2  & LLaMA-2-Chat-7B & \textcolor{deepred}{\textbf{56.9}} & \textcolor{blue}{\textbf{47.7}} & -- & 60.3 & 30.3 & -- & 51.9 & -- & -- & -- & -- & -- & -- & -- & -- & -- & \textcolor{blue}{\textbf{58.2}} & \textcolor{blue}{\textbf{60.6}} \\
MiniGPT-v2-Chat  & LLaMA-2-Chat-7B & \textcolor{blue}{\textbf{55.9}} & \textcolor{deepred}{\textbf{49.4}} & -- & 58.8 & 42.4 & -- & 52.3 & -- & -- & -- & -- & -- & -- & -- & -- & -- & \textcolor{deepred}{\textbf{59.5}} & \textcolor{deepred}{\textbf{63.3}} \\
VILA-7B  & LLaMA-2-7B & -- & -- & 79.9$^*$  & {62.3}$^*$ & {57.8} & 68.2 & {64.4} & \textcolor{blue}{\textbf{85.5}} & {1533.0}& -- & {68.9} & 61.7 & 61.1 & 69.7 & 34.9 & -- & -- & -- \\
VILA-13B  & LLaMA-2-13B & -- & -- & {\textcolor{deepred}{\textbf{80.8}}}$^*$ & {\textcolor{deepred}{\textbf{63.3}}}$^*$ & \textcolor{blue}{\textbf{60.6}} & \textcolor{deepred}{\textbf{73.7}} & \textcolor{deepred}{\textbf{66.6}} & 84.2 & \textcolor{deepred}{\textbf{1570.1}}& -- & \textcolor{blue}{\textbf{70.3}} & \textcolor{blue}{\textbf{64.3}} & \textcolor{blue}{\textbf{62.8}} & \textcolor{blue}{\textbf{73.0}} & \textcolor{blue}{\textbf{38.8}} & -- & -- & -- \\
\quad +ShareGPT4V & LLaMA-2-13B & -- & -- &\textcolor{blue}{\textbf{80.6}}$^*$ & \textcolor{blue}{\textbf{63.2}}$^*$ & \textcolor{deepred}{\textbf{62.4}} & \textcolor{blue}{\textbf{73.1}} & \textcolor{blue}{\textbf{65.3}} & 84.8 & 1556.5 & -- & \textcolor{deepred}{\textbf{70.8}} & \textcolor{deepred}{\textbf{65.4}} & 61.4 & \textcolor{deepred}{\textbf{78.4}} & \textcolor{deepred}{\textbf{45.7}} & -- & -- & -- \\
\bottomrule
\end{tabular}
}
\caption{Comparison of mainstream MM-LLMs on $18$ VL benchmarks. The \textcolor{deepred}{\textbf{red}} denotes the highest result, and the \textcolor{blue}{\textbf{blue}} denotes the second highest result. $^{\ddagger}$ indicates ShareGPT4V's~\cite{chen2023sharegpt4v} re-implemented test results, which are missed in benchmarks or origin papers. $^*$ indicates that training images are observed during training. 
}
\label{tab:benchmarks}
\end{table*}

\paragraph{Trends in Existing MM-LLMs:} \textbf{(1)} Progressing from a dedicated emphasis on MM understanding to the generation of specific modalities and further evolving into any-to-any modality conversion (\emph{e.g.}, MiniGPT-4 $\to$ MiniGPT-5 $\to$ NExT-GPT); \textbf{(2)} Advancing from MM PT to SFT and then to RLHF, the training pipeline undergoes continuous refinement, striving to better align with human intent and enhance the model's conversational interaction capabilities (\emph{e.g.}, BLIP-2 $\to$ InstructBLIP $\to$ DRESS); \textbf{(3)} Embracing Diversified Modal Extensions (\emph{e.g.}, BLIP-2 $\to$ X-LLM and InstructBLIP $\to$ X-InstructBLIP); \textbf{(4)} Incorporating a Higher-Quality Training Dataset (\emph{e.g.}, LLaVA $\to$ LLaVA-1.5); \textbf{(5)} Adopting a More Efficient Model Architecture, transitioning from complex Q- and P-Former input projector modules in BLIP-2 and DLP to a simpler yet effective linear projector in VILA.

\section{Benchmarks and Performance}\label{benchmark} 

To provide a comprehensive performance comparison, we have compiled a table featuring major MM-LLMs across $18$ Vision-Language (VL) benchmarks, as reported in various papers~\cite{DBLP:conf/icml/0008LSH23,chen2023minigpt,chen2023sharegpt4v,lin2023vila}. This information is presented in Table~\ref{tab:benchmarks}, with detailed descriptions of these benchmarks available in Appendix~\ref{ap:benchmark}. Given the numerous benchmarks available, we focus on evaluating and comparing different MM-LLMs based on OKVQA, IconVQA, VQA$^\text{v2}$, and GQA. 

OKVQA includes questions requiring reasoning with a variety of knowledge types such as commonsense, world knowledge, and visual knowledge. MiniGPT-v2 and MiniGPT-v2-chat perform best in this benchmark, showcasing their outstanding reasoning abilities. IconVQA emphasizes the importance of abstract diagram comprehension and holistic cognitive reasoning in real-world diagram-based word problems, requiring both perceptual acumen and versatile cognitive reasoning. MiniGPT-v2 and MiniGPT-v2-chat also excel in this benchmark, highlighting their exceptional perception and cognitive reasoning capabilities. VQA$^\text{v2}$ is a more balanced VQA dataset where each question is paired with a series of images. VILA-13B performs best in this benchmark, demonstrating its superior ability to comprehend multimodal information and its resistance to language biases in the knowledge it acquires. GQA is a VQA dataset focusing on image scene graphs, offering impartial compositional questions derived from real-world images. Each question is associated with a structured representation of its meaning and the detailed logical steps required to answer it. LLaVA-1.5 and VILA-7B perform best in this benchmark, illustrating their excellent reasoning abilities in this domain. 

Following this, we will outline training recipes that enhance the effectiveness of MM-LLMs, drawing insights from SOTA models.

\paragraph{Training Recipes}
\textbf{Firstly}, higher image resolution can incorporate more visual details for the model, benefiting tasks that require fine-grained details. For example, LLaVA-1.5 and VILA employ a resolution of $336\times336$, while Qwen-VL and MiniGPT-v2 utilize $448\times448$. However, higher resolutions lead to longer token sequences, incurring additional training and inference costs. MiniGPT-v2 addresses this by concatenating $4$ adjacent visual tokens in the embedding space to reduce length. Recently, Monkey~\cite{li2023monkey} proposed a solution to enhance the resolution of input images without retraining a high-resolution visual encoder, utilizing only a low-resolution visual encoder, supporting resolutions up to $1300\times800$. To enhance the understanding of rich-text images, tables, and document content, DocPedia~\cite{feng2023docpedia} introduced a method to increase the visual encoder resolution to $2560\times2560$, overcoming the limitations of poorly performing low resolutions in open-sourced ViT. 
\textbf{Secondly}, the incorporation of high-quality SFT data can significantly improve performance in specific tasks, as evidenced by the addition of ShareGPT4V data to LLaVA-1.5 and VILA-13B, as shown in Table~\ref{tab:benchmarks}.
\textbf{Moreover}, VILA reveals several key findings: (1) Performing PEFT on the LLM Backbone promotes deep embedding alignment, crucial for ICL; (2) Interleaved Image-Text data proves beneficial, whereas Image-Text pairs alone are sub-optimal; (3) Re-blending text-only instruction data (\emph{e.g.}, unnatural instruction~\cite{honovich2022unnatural}) with image-text data during SFT not only addresses the degradation of text-only tasks but also enhances VL task accuracy.

\section{Future Directions}\label{future}

In this section, we explore promising future directions for MM-LLMs across the following aspects:

\paragraph{More General and Intelligent Models} We can enhance the MM-LLMs' strength from the following four key avenues: \textbf{(1) Expanding Modalities}: Current MM-LLMs mainly support the following modalities: image, video, audio, 3D, and text. However, the real world involves a broader range of modalities. Extending MM-LLMs to accommodate additional modalities (\emph{e.g.}, web pages, heat maps, and figures\&tables) will increase the model's versatility, making it more universally applicable;
\textbf{(2) Diversifying LLMs}: Incorporating various types and sizes of LLMs provides practitioners with the flexibility to select the most appropriate one based on their specific requirements;
\textbf{(3) Improving MM IT Dataset Quality}: Current MM IT datasets have ample room for improvement and expansion. 
Diversifying the range of instructions can enhance the effectiveness of MM-LLMs in understanding and executing user commands;
\textbf{(4) Strengthening MM Generation Capabilities}: Most current MM-LLMs are predominantly oriented towards MM understanding. Although some models have incorporated MM generation capabilities, the quality of generated responses may be constrained by the capacities of the LDMs. 
Exploring the integration of retrieval-based approaches~\cite{asai2023retrieval,gao2023retrieval,kang2024c} holds significant promise in complementing the generative process, enhancing the overall performance of the model.

\paragraph{More Challenging Benchmarks}

Existing benchmarks may not sufficiently challenge the capabilities of MM-LLMs, as many datasets have appeared to varying degrees in the PT or IT sets. This implies that the models might have already learned these tasks during training. Moreover, current benchmarks predominantly focus on the VL sub-field. Therefore, it is crucial for the development of MM-LLMs to construct a more challenging, larger-scale benchmark that includes additional modalities and employs a unified evaluation standard. For instance, GOAT-Bench~\cite{lin2024goat} is designed to assess the capability of various MM-LLMs in discerning and responding to nuanced aspects of social abuse depicted in memes. MMCode~\cite{Li2024MMCodeEM} evaluates the algorithmic problem-solving skills of MM-LLMs in visually rich contexts. DecodingTrust~\cite{wang2024decodingtrust} measures the trustworthiness of MM-LLMs. MathVista~\cite{lu2023mathvista} evaluates the mathematical reasoning ability of MM-LLMs within visual contexts, while GeoEval~\cite{zhang2024geoeval,li2024lans,song2024fmint} assesses their proficiency in tackling geometry math problems. Moreover, MMMU~\cite{yue2023mmmu} and CMMMU~\cite{zhang2024cmmmu} have respectively introduced English and Chinese versions of a comprehensive multi-discipline MM understanding and reasoning benchmark for expert artificial general intelligence. Additionally,~\citeauthor{fan2024muffin} have challenged MM-LLMs with multipanel VQA, and BenchLMM~\cite{cai2023benchlmm} benchmarks the cross-style visual capability of MM-LLMs. Furthermore,~\citeauthor{liu2023hidden} have conducted an in-depth study on the optical character recognition capabilities of MM-LLMs. These efforts highlight the need for more sophisticated and diverse benchmarks to truly gauge the advanced capabilities of MM-LLMs.

\paragraph{Mobile/Lightweight Deployment}
To deploy MM-LLMs on resource-constrained platforms and achieve optimal performance meanwhile, such as low-power mobile and IoT devices, lightweight implementations are of paramount importance. A notable advancement in this realm is MobileVLM~\cite{chu2023mobilevlm}. This approach strategically downscales LLaMA, allowing for seamless off-the-shelf deployment. MobileVLM further introduces a lightweight downsample projector, consisting of fewer than $20$ million parameters, contributing to improved computational speed. Recently, there have been many similar studies on lightweighting MM-LLMs, achieving efficient computation and inference with comparable performance or minimal loss, including TinyGPT-V~\cite{yuan2023tinygpt}, Vary-toy~\cite{wei2024small}, Mobile-Agent~\cite{wang2024mobile}, MoE-LLaVA~\cite{lin2024moe}, and MobileVLM V2~\cite{chu2024mobilevlm2}.
Nevertheless, this avenue necessitates additional exploration for further advancements in development.

\paragraph{Embodied Intelligence} The embodied intelligence aims to replicate human-like perception and interaction with the surroundings by effectively understanding the environment, recognizing pertinent objects, assessing their spatial relationships, and devising a comprehensive task plan~\cite{firoozi2023foundation}. Embodied AI tasks, such as embodied planning, embodied visual question answering, and embodied control, equip robots to autonomously implement extended plans by leveraging real-time observations. Some typical works in this area are PaLM-E~\cite{driess2023palm} and EmbodiedGPT~\cite{mu2023embodiedgpt}. PaLM-E introduces a multi-embodiment agent through the training of a MM-LLM. Beyond functioning solely as an embodied decision maker, PaLM-E also demonstrates proficiency in handling general VL tasks. EmbodiedGPT introduces an economically efficient method characterized by a CoT approach, enhancing the capability of embodied agents to engage with the real world and establishing a closed loop that connects high-level planning with low-level control. While MM-LLM-based Embodied Intelligence has made advancements in integrating with robots, further exploration is needed to enhance the autonomy of robots.

\paragraph{Continual Learning} 

Due to the large training costs associated with their massive scale, MM-LLMs are not amenable to frequent re-training. However, updates are necessary to endow MM-LLMs with new skills and keep them up-to-date with rapidly evolving human knowledge~\cite{wu2024continual}. Thus, Continual Learning (CL) is needed to make the model flexible enough to efficiently and continually leverage emerging data while avoiding the substantial cost of retraining MM-LLMs.
CL for MM-LLMs can be classified into two stages: continual PT and continual IT. Recently, a continual MM IT benchmark has been proposed to continuously fine-tune MM-LLMs for new MM tasks while maintaining superior performance on tasks learned during the original MM IT stage~\cite{he2023continual}. It introduces two primary challenges: (1) catastrophic forgetting, where models forget previous knowledge when learning new tasks~\cite{robins1995catastrophic,mccloskey1989catastrophic,goodfellow2013empirical,zhang2023decomposing,zhang2023task,zhang2023continual,zheng2023learn}, and (2) negative forward transfer, indicating that the performance of unseen tasks declines when learning new ones~\cite{zheng2024beyond,dong2022federated_FCIL,dong2022federated_FCIL_PAMI,dong2023federated,dong2023heterogeneous}.

\paragraph{Mitigating Hallucination} Hallucinations entail generating textual descriptions of nonexistent objects without visual cues, which manifest in diverse categories~\cite{liu2024survey} such as misjudgments and inaccuracies in descriptions. The origins of these hallucinations are multifaceted~\cite{liu2024survey}, including biases and annotation errors in training data. Additionally, Skip $\textbackslash n$~\cite{han2024skip} highlights semantic drift biases associated with paragraph separators, which can induce hallucinations when deliberately inserted. Current methods to mitigate these hallucinations involve leveraging self-feedback as visual cues~\cite{lee2023volcano}. However, challenges persist, necessitating nuanced discernment between accurate and hallucinatory outputs, as well as advancements in training methodologies to enhance output reliability.

\paragraph{Biases and Ethical Considerations} 
Despite the strengths of MM-LLMs, ensuring their safe and efficient application remains crucial. Information generated by MM-LLMs can perpetuate stereotypes and cause harm to vulnerable populations. Since MM-LLMs learn from patterns in MM training data, they can reproduce biases present in these data, potentially leading to representational harm.
To address this, we can develop new benchmarks specifically designed to evaluate biases in MM-LLMs \cite{luo2024fairclip}. Additionally, designing more effective and fine-grained alignment methods is essential. For instance, using RLHF can help calibrate MM-LLMs to produce answers that align with human values and desires \cite{li2024red}.

\section{Conclusion}\label{conclusion}
In this paper, we have presented a comprehensive survey of MM-LLMs with a focus on recent advancements. Initially, we categorize the model architecture into five components, providing a detailed overview of general design formulations and training pipelines. Subsequently, we introduce various SOTA MM-LLMs, each distinguished by its specific formulations. Our survey also sheds light on their capabilities across diverse MM benchmarks and envisions future developments in this rapidly evolving field. We hope this survey can provide insights for researchers, contributing to the ongoing advancements in the MM-LLMs domain.

\section*{Social Impact}

MM-LLMs have the potential to impact society. They can enhance accessibility for individuals with disabilities by improved voice recognition and visual aids, fostering equal access to information. In education, MM-LLMs can revolutionize learning with more interactive experiences, catering to diverse learning styles. In media, they can create more engaging content, enriching the consumer experience. However, the widespread adoption of MM-LLMs also poses risks. Privacy concerns arise from the vast training data, raising issues of data security and user privacy. There is also a risk of exacerbating biases in AI algorithms, as biases in training data can lead to biased outputs. Additionally, the automation of tasks traditionally performed by humans could lead to job displacement, necessitating proactive measures to mitigate potential negative impacts on employment. Overall, while MM-LLMs offer promising opportunities, it is essential to address these challenges to ensure their responsible and equitable deployment.

\newpage

\section*{Acknowledgments}

We express our gratitude to the anonymous reviewers for their valuable and insightful comments. This work was supported by JSPS KAKENHI Grant Number JP23K28144.

\section*{Limitations}

In this paper, we embark on a comprehensive exploration of the current MM-LLMs landscape, presenting a synthesis from diverse perspectives enriched by our insights. Acknowledging the dynamic nature of this field, it is plausible that certain aspects may have eluded our scrutiny, and recent advances might not be entirely encapsulated. To tackle this inherent challenge, we've established a dedicated website for real-time tracking, using crowdsourcing to capture the latest advancements. Our goal is for this platform to evolve into a continuous source of contributions propelling ongoing development in the field.
Given the constraints of page limits, we are unable to delve into all technical details and have provided concise overviews of the core contributions of mainstream MM-LLMs.
Looking ahead, we commit to vigilant monitoring and continual enhancement of relevant details on our website (\href{https://mm-llms.github.io}{\textcolor{blue}{https://mm-llms.github.io}}), incorporating fresh insights as they emerge.

\bibliography{anthology,custom}

\newpage

\appendix

\section{Related Surveys}
\label{ap:rw}
Prior to the emergence of LLMs, several surveys on traditional MM PT have been conducted~\cite{ruan2022survey,DBLP:conf/ijcai/DuLLZ22,DBLP:conf/ijcai/LongCHY22,chen2023vlp}. Most of these models entail a substantial computational cost during the PT phase, attributable to end-to-end training using large-scale models and datasets. As a consequence of not incorporating LLMs, these models suffer from deficiencies in instruction following, ICL, CoT, and interactive capabilities. Moreover, the training pipeline solely encompasses the PT phase without the inclusion of an IT stage. 

In recent times, several surveys have emerged on MM-LLMs. \citeauthor{yin2023survey} and~\citeauthor{wu2023multimodal} exclusively delve into early VL understanding models. \citeauthor{huang2023visual} place a primary emphasis on visual IT, while~\citeauthor{song2023bridge} focus on modal alignment methods. Lastly,~\citeauthor{cui2024survey} provide a comprehensive review of the applications of MM-LLMs within the realm of autonomous driving.

Compared with their works, the main distinctions are outlined as follows: 
\begin{itemize}
\item We have comprehensively covered nearly all MM-LLMs over the past year, totaling around $120$ or more, including not only understanding models but also generative models. Our coverage extends beyond VL modalities to encompass various modes such as audio and 3D point cloud; 
\item To offer readers a comprehensive understanding of MM-LLMs, we have introduced a general model architecture that incorporates any-to-any modality transformations, offering a detailed overview of the functional roles and implementation choices for each component; 

\item We have summarized the developmental trends of existing MM-LLMs and provided some training recipes that can enhance effectiveness; 

\item We have established an open-source website (\href{https://mm-llms.github.io}{\textcolor{blue}{https://mm-llms.github.io}}) for MM-LLMs researchers, supporting crowdsourced updates and aiming to facilitate collaboration in the MM-LLMs field. We anticipate that this survey will illuminate future research in the MM-LLMs domain.

\end{itemize}

\section{Modality Encoder}\label{ap:encoder}

In the following, we provide a brief introduction to some mainstream modality encoders.

\subsection{Visual Modality }

\paragraph{NFNet-F6}~\cite{brock2021high} is a normalizer-free ResNet~\cite{he2016deep}, showcasing an adaptive gradient clipping that allows training on extensively augmented datasets while achieving SOTA levels of image recognition.
\paragraph{ViT}~\cite{dosovitskiy2020image} applies the Transformer~\cite{vaswani2017attention} to images by first dividing the image into patches. It then undergoes linear projection to flatten the patches, followed by encoding via Transformer blocks.
\paragraph{CLIP ViT}~\cite{radford2021learning} builds connections between text and images, comprising a ViT and a text encoder. With a vast amount of text-image pairs, it optimizes ViT by contrastive learning, treating paired text and images as positive samples and others as negative ones.
\paragraph{Eva-CLIP ViT}~\cite{fang2023eva} stabilizes the training and optimization process of the massive CLIP, offering new directions in expanding and accelerating the expensive training of MM base models.

\subsection{Audio Modality}

\paragraph{C-Former}~\cite{chen2023x} employs the CIF~\cite{dong2020cif,DBLP:conf/ijcai/ZhangZJW022,han2022improving,han2023knowledge} for sequence transduction and a Transformer to extract audio features.
\paragraph{HuBERT}~\cite{hsu2021hubert} is a self-supervised speech representation learning framework based on BERT~\cite{kenton2019bert}, achieved by the masked prediction of discrete hidden units. 
It has the capability to convert continuous speech signals into a sequence of discrete units.
\paragraph{BEATs}~\cite{DBLP:conf/icml/ChenW00T0CYW23} is an iterative audio pre-training framework designed to learn Bidirectional Encoder representations from Audio Transformers.

\section{Mainstream PEFT Methods }
\label{ap:peft}

PEFT entails maintaining the pre-trained LLM in a frozen state while adjusting a small number of additional trainable parameters. In the following section, we revisit several representative PEFT methods, where $\bm{x}$ and $\bm{h}$ represent the input and output of the original module, and $\bm{h}'$ signifies the output of this module when attached with PEFT.

\paragraph{Prefix-tuning}\cite{li2021prefix,lester2021power} involves the addition of learnable tokens to the keys and values of the attention module. This process is formulated as follows:
\begin{equation}
  \begin{aligned}
    \mathbf{h}' = \operatorname{Attn}\left(
    \mathbf{x} \mathbf{W}_q,
    [\mathbf{P}_k, \mathbf{x} \mathbf{W}_k],
    [\mathbf{P}_v, \mathbf{x} \mathbf{W}_v]
    \right)\text{,}
  \end{aligned}
\end{equation}
with $\mathbf{P}_k, \mathbf{P}_v \in \mathbb{R}^{l \times d}$ representing two sets of prefix tokens. $[\cdot, \cdot]$ denotes concatenation, and $\operatorname{Attn}$ is defined as:
$$
  \operatorname{Attn}\left(\mathbf{Q}, \mathbf{K}, \mathbf{V} \right) := \operatorname{softmax}\left(\frac{\mathbf{Q} \mathbf{K}^T}{\sqrt{d}}\right) \mathbf{V}\text{.}
$$

\paragraph{Adapter}\cite{houlsby2019parameter,he2021towards,rebuffi2017learning,zhang2020side} is typically a residual block consisting of a down-projection matrix $\mathbf{A}$, a nonlinear activation function $\sigma(\cdot)$, and an up-projection matrix $\mathbf{B}$. It can be inserted into any layer of the pre-trained LLM, formulated as follows:
\begin{equation}
  \mathbf{h}' = \mathbf{h} + \sigma(\textbf{x}\mathbf{A})\mathbf{B}.
\end{equation}

\paragraph{LoRA}~\cite{hu2021lora} is the most commonly used PEFT method. It assumes that the change in parameters occurs within a low-rank space. Given a pre-trained matrix $\bm{W}\in\mathbb{R}^{c\times d}$, LoRA learns an incremental update $\Delta\bm{W}$ and decomposes $\Delta\bm{W}$ into a matrix multiplication between two low-rank matrices $\bm{A}\in\mathbb{R}^{c\times r}$ and $\bm{B}\in\mathbb{R}^{r\times d}$, where $r$ $\ll$ min$(c, d)$. LoRA follows the forward process as outlined below:
\begin{equation}
    \bm{h} = \bm{W}\bm{x} + \Delta\bm{W}\bm{x}=  \bm{W}\bm{x} + \bm{AB}\bm{x}\text{.}
\end{equation}

QLoRA~\cite{dettmers2023qlora} is a quantized LoRA. The underlying principle of QLoRA includes the quantization of pre-trained weights to $4$ bits, followed by the execution of PEFT using LoRA.

\paragraph{LayerNorm tuning}~\cite{zhao2023tuning} presents an efficient strategy to transform LLMs into MM-LLMs, which tunes LayerNorm in attention block yielding strong MM performance compared with full parameter finetuning or LoRA. 

In addition to the aforementioned PEFT methods, there are several others, including P-tuning~\cite{liu2022p}, P-tuning v2~\cite{liu2021p}, AdaptBias~\cite{fu2022adapterbias}, Compacter~\cite{karimi2021compacter}, AdapterFormer~\cite{chen2022adaptformer}, XTuner~\cite{2023xtuner}, P-LoRA~\cite{dong2024internlm}, MoLE~\cite{chen2024llava}, and Delta-LoRA~\cite{zi2023delta}.

\section{Representative LLMs}
\label{ap:model}

The representative LLM Backbones in existing MM-LLMs research are as follows:

    \paragraph{Flan-T5}~\cite{chung2022scaling} investigates IT for T5~\cite{raffel2020exploring}, an encoder-decoder architecture using unified text-to-text training for all natural language processing issues, exhibiting robust zero-shot and CoT capabilities.

\paragraph{ChatGLM} is a Chinese-English bilingual dialogue model,\footnote{\url{https://github.com/THUDM/ChatGLM-6B}} optimized by an auto-regressive mask infilling objective. It is based on the GLM~\cite{du2022glm,zeng2022glm} architecture, optimized for Chinese question answering and dialogues.

\paragraph{InternLM}~\cite{team2023internlm} is a multilingual trillion-parameter foundation model trained on over a trillion tokens of data. Based on this foundation, the model utilizes high-quality human-annotated dialogue data combined with RLHF to respond to complex instructions during human interactions, exhibiting responses that align with human ethics and values.

\paragraph{UL2}~\cite{tay2022ul2} is an encoder-decoder model trained utilizing a mixture of denoisers objectives, surpassing T5 on numerous benchmarks.

\paragraph{Qwen}~\cite{bai2023qwen} is trained on large-scale and diverse datasets, with a primary focus on Chinese and English. It employs SFT and RLHF techniques for alignment, resulting in dialogue models like Qwen-Chat.

\paragraph{Chinchilla}~\cite{hoffmann2022training} is a causal decoder, trained on extensive text data. It posits that model size should double for every doubling of training tokens.

\paragraph{OPT}~\cite{zhang2022opt} is a GPT-3~\cite{brown2020language} clone, striving to release an open-source model that replicates the performance of GPT-3.

\paragraph{PaLM}~\cite{chowdhery2023palm} is a causal decoder structure with parallel attention and feed-forward layers, enabling training speeds up to $15$ times faster. Notable changes contain RoPE embeddings, SwiGLU activation, multi-query attention, and etc.

\paragraph{LLaMA}~\cite{touvron2023llama} comprises decoder-only models with efficient causal attention. 
    
\paragraph{LLaMA-2}~\cite{touvron2023llama2} focuses on fine-tuning a superior and safer LLaMA-2-Chat model for conversation generation, incorporating 40\% more training data with grouped-query attention and a larger context length.

\paragraph{Vicuna}~\cite{vicuna2023} is a model built on top of LLaMA, utilizing user dialogue data obtained from \url{ShareGPT.com} and trained by SFT.

\section{SOTA MM-LLMs}\label{ap:sota}

In the following, we will provide a brief introduction to the core contributions of some representative MM-LLMs.

\paragraph{Flamingo}~\cite{alayrac2022flamingo} represents a series of Visual Language (VL) Models designed for processing interleaved visual data and text, generating free-form text as the output.
\paragraph{BLIP-2}~\cite{DBLP:conf/icml/0008LSH23} introduces a more resource-efficient framework, comprising the lightweight Q-Former to bridge modality gaps and the utilization of frozen LLMs. Leveraging LLMs, BLIP-2 can be guided for zero-shot image-to-text generation using natural language prompts.
\paragraph{LLaVA}~\cite{liu2023llava} pioneers the transfer of IT techniques to the MM domain. Addressing data scarcity, LLaVA introduces a novel open-source MM instruction-following dataset created using ChatGPT/GPT-4, alongside the MM instruction-following benchmark, LLaVA-Bench.
\paragraph{MiniGPT-4}~\cite{zhu2023minigpt} proposes a streamlined approach where training only one linear layer aligns the pre-trained vision encoder with the LLM. This efficient method enables the replication of the exhibited capabilities of GPT-4.
\paragraph{mPLUG-Owl}~\cite{ye2023mplug} presents a novel modularized training framework for MM-LLMs, incorporating the visual context. To assess different models' performance in MM tasks, the framework includes an instructional evaluation dataset called OwlEval.
\paragraph{X-LLM}~\cite{chen2023x} is expanded to various modalities, including audio, and demonstrates strong scalability. Leveraging the language transferability of the Q-Former, X-LLM is successfully applied in the context of Sino-Tibetan Chinese.
\paragraph{VideoChat}~\cite{li2023videochat} pioneers an efficient chat-centric MM-LLM for video understanding dialogue, setting standards for future research in this domain and offering protocols for both academia and industry.
\paragraph{InstructBLIP}~\cite{DBLP:journals/corr/abs-2305-06500} is trained based on the pre-trained BLIP-2 model, updating only the Q-Former during MM IT. By introducing instruction-aware visual feature extraction and corresponding instructions, the model enables the extraction of flexible and diverse features.
\paragraph{PandaGPT}~\cite{su2023pandagpt} is a pioneering general-purpose model with the capability to comprehend and act upon instructions across $6$ different modalities: text, image/video, audio, thermal, depth, and inertial measurement units.
\paragraph{PaLI-X}~\cite{chen2023pali} is trained using mixed VL objectives and unimodal objectives, including prefix completion and masked-token completion. This approach proves effective for both downstream task results and achieving the Pareto frontier in the fine-tuning setting.
\paragraph{Video-LLaMA}~\cite{DBLP:conf/emnlp/ZhangLB23} introduces a multi-branch cross-modal PT framework, enabling LLMs to simultaneously process the vision and audio content of a given video while engaging in conversations with humans. This framework aligns vision with language as well as audio with language.
\paragraph{Video-ChatGPT}~\cite{maaz2023video} is a model specifically designed for video conversations, capable of generating discussions about videos by integrating spatiotemporal vision representations.
\paragraph{Shikra}~\cite{chen2023shikra} introduces a simple and unified pre-trained MM-LLM tailored for Referential Dialogue, a task involving discussions about regions and objects in images. This model demonstrates commendable generalization ability, effectively handling unseen settings.
\paragraph{DLP}~\cite{jian2023bootstrapping} proposes the P-Former to predict the ideal prompt, trained on a dataset of single-modal sentences. This showcases the feasibility of single-modal training to enhance MM learning.
\paragraph{BuboGPT}~\cite{zhao2023bubogpt} is a model constructed by learning a shared semantic space for a comprehensive understanding of MM content. It explores fine-grained relationships among different modalities such as image, text, and audio.
\paragraph{ChatSpot}~\cite{zhao2023chatspot} introduces a simple yet potent method for finely adjusting precise referring instructions for MM-LLM, facilitating fine-grained interactions. The incorporation of precise referring instructions, consisting of image- and region-level instructions, enhances the integration of multi-grained VL task descriptions.
\paragraph{Qwen-VL}~\cite{Qwen} is a multi-lingual MM-LLM that supports both English and Chinese. Qwen-VL also allows the input of multiple images during the training phase, improving its ability to understand the vision context.
\paragraph{NExT-GPT}~\cite{wu2023next} is an end-to-end, general-purpose any-to-any MM-LLM that supports the free input and output of image, video, audio, and text. It employs a lightweight alignment strategy, utilizing LLM-centric alignment in the encoding phase and instruction-following alignment in the decoding phase.
\paragraph{MiniGPT-5}~\cite{zheng2023minigpt} is an MM-LLM integrated with inversion to generative vokens and integration with Stable Diffusion. It excels in performing interleaved VL outputs for MM generation. The inclusion of classifier-free guidance during the training phase enhances the quality of generation.

\paragraph{LLaVA-1.5}~\cite{liu2023improved} reports simple modifications to the LLaVA framework, including applying an MLP projection and introducing VQA data tailored for academic tasks, along with simple response formatting prompts. These adjustments result in enhanced capabilities for MM understanding. 

\paragraph{MiniGPT-v2}~\cite{chen2023minigpt} is an MM-LLM designed as a unified interface for diverse VL multi-task learning. To create a single model proficient in handling multiple VL tasks, identifiers are incorporated for each task during both training and inference. This facilitates clear task distinction, ultimately enhancing learning efficiency.

\paragraph{CogVLM}~\cite{wang2023cogvlm} is an open-source MM-LLM that bridges the gap between modalities via a trainable visual expert module within the attention and feedforward layers. This allows for a deep fusion of MM features without compromising performance on NLP downstream tasks.

\paragraph{DRESS}~\cite{chen2023dress} introduces a method using natural language feedback to enhance alignment with human preferences. DRESS extends the conditional reinforcement learning algorithm to integrate non-differentiable natural language feedback, training the model to generate appropriate responses based on feedback.

\paragraph{X-InstructBLIP}~\cite{panagopoulou2023x} introduces a cross-modal framework with instruction-aware representations, scalable enough to empower LLMs to handle diverse tasks across multiple modalities, including image/video, audio, and 3D. Notably, it achieves this without the need for modality-specific PT.

\paragraph{CoDi-2}~\cite{tang2023codi2} is a MM generation model excelling in modality-interleaved instruction following, in-context generation, and user-model interaction by multi-turn conversations. It enhances CoDi~\cite{tang2023codi} to process intricate modality-interleaved inputs and instructions, generating latent features autoregressively.

\paragraph{VILA}~\cite{lin2023vila} outperforms in vision tasks and shows remarkable reasoning ability while maintaining text-only capabilities. It achieves this by harnessing the full capabilities of LLM learning, using the interleaved attributes of image-text pairs, and implementing meticulous text data re-blending.

\begin{table*}[tbp]
\centering
\resizebox{1.0\linewidth}{!}{
\begin{tabular}{l c c c c}
\toprule
\textbf{Dataset Name} &  \textbf{X Modality} & \textbf{\#.X} & \textbf{\#.T} & \textbf{\#.X-T}\\ 
\midrule
ALIGN~\cite{jia2021scaling} & Image & 1.8B & 1.8B & 1.8B \\
LTIP~\cite{alayrac2022flamingo} & Image & 312M & 312M & 312M \\
MS-COCO~\cite{lin2014microsoft} & Image & 124K & 620K & 620K \\
Visual Genome~\cite{krishna2017visual} & Image & 108K & 4.5M & 4.5M\\
CC3M~\cite{sharma2018conceptual} & Image & 3.3M & 3.3M & 3.3M \\
CC12M~\cite{changpinyo2021conceptual} & Image & 12.4M & 12.4M & 12.4M \\
SBU~\cite{ordonez2011im2text} & Image & 1M & 1M & 1M \\
LAION-5B~\cite{schuhmann2022laion} & Image & 5.9B & 5.9B & 5.9B \\
LAION-400M~\cite{schuhmann2021laion} & Image & 400M & 400M & 400M \\
LAION-en~\cite{schuhmann2022laion} & Image & 2.3B & 2.3B & 2.3B \\
LAION-zh~\cite{schuhmann2022laion} & Image & 142M & 142M & 142M \\
LAION-COCO~\cite{laioncoco} & Image & 600M & 600M & 600M \\
Flickr30k~\cite{young2014image} & Image & 31K & 158K & 158K \\
AI Challenger Captions~\cite{wu2017ai} & Image & 300K & 1.5M & 1.5M \\
COYO~\cite{coyo} & Image & 747M & 747M & 747M \\
Wukong~\cite{gu2022wukong} & Image & 101M & 101M &101M \\
COCO Caption~\cite{chen2015microsoft} & Image & 164K  & 1M  & 1M \\
WebLI~\cite{chen2022pali} & Image & 10B  & 12B   & 12B  \\
Episodic WebLI~\cite{chen2023pali} & Image & 400M  &  400M  & 400M  \\
CC595k~\cite{liu2023llava} & Image & 595K  &  595K  & 595K  \\
RefCOCO~\cite{kazemzadeh2014referitgame} & Image & 20K  & 142K & 142K \\
RefCOCO+~\cite{yu2016modeling} & Image & 20K  & 142K & 142K \\
Visual-7W~\cite{zhu2016visual7w} & Image & 47.3K  & 328K & 328K \\
OCR-VQA~\cite{mishra2019ocr} & Image & 207K  & 1M & 1M \\
ST-VQA~\cite{biten2022latr} & Image & 23K  & 32K & 32K \\
DocVQA~\cite{mathew2021docvqa} & Image & 12K & 50K & 50K \\
TextVQA~\cite{singh2019towards} & Image & 28.4K & 45.3K & 45.3K \\
DataComp~\cite{gadre2023datacomp} & Image & 1.4B & 1.4B & 1.4B \\
GQA~\cite{hudson2019gqa} & Image & 113K & 22M & 22M \\
VGQA~\cite{krishna2017visual} & Image & 108K & 1.7M & 1.7M \\
VQA$^\text{v2}$~\cite{goyal2017making} & Image & 265K & 1.4M & 1.4M \\
DVQA~\cite{kafle2018dvqa} & Image & 300K & 3.5M & 3.5M \\
OK-VQA~\cite{schwenk2022okvqa} & Image & 14K & 14K & 14K \\
A-OKVQA~\cite{schwenk2022okvqa} & Image & 23.7K & 24.9K & 24.9K \\
Text Captions~\cite{sidorov2020textcaps} & Image & 28K & 145K & 145K \\
Multimodal Arxiv~\cite{li2024multimodal} & Image & 32K & 16.6K & 16.6K \\
M3W (Interleaved)~\cite{alayrac2022flamingo} & Image & 185M & 182GB & 43.3M (Instances) \\
MMC4 (Interleaved)~\cite{zhu2023multimodal} & Image & 571M & 43B & 101.2M (Instances) \\
Obelics (Interleaved)~\cite{laurenccon2023obelics} & Image & 353M & 115M & 141M (Instances) \\
MSRVTT~\cite{xu2016msr} & Video & 10K & 200K & 200K \\
WebVid~\cite{bain2021frozen} & Video & 10M & 10M & 10M \\
VTP~\cite{alayrac2022flamingo} & Video & 27M & 27M & 27M \\
AISHELL-1~\cite{chen2023x} & Audio & -- & -- & 128K \\
AISHELL-2~\cite{chen2023x} & Audio & -- & -- & 1M \\

WaveCaps~\cite{mei2023wavcaps} & Audio & 403K & 403K & 403K \\

VSDial-CN~\cite{ni2023vilas} & Image, Audio & 120K (Image), 1.2M(Audio) & 120K & 1.2M \\

\bottomrule
\end{tabular}
}
\caption{The statistics for MM PT datasets. \textbf{\#.X} represents the quantity of X, \textbf{\#.T} represents the quantity of Text, and \textbf{\#.X-T} represents the quantity of X-Text pairs, where X can be Image, Video, or Audio.}

\label{PT_corpus}
\end{table*}

\begin{table*}[htbp]
\centering
\resizebox{1.0\linewidth}{!}{
\begin{tabular}{l c c c c c c c c}
\toprule
\textbf{Dataset Name} & \textbf{Type} &	\textbf{I$\to$O} & \textbf{Source} & \textbf{Method} & \textbf{Multi-Turn} &	\textbf{\#.I/V/A}	& \textbf{\#.Dialog Turn} &\textbf{\#.Instance}\\
\midrule
MiniGPT-4's IT~\cite{zhu2023minigpt}& SFT & I+T$\to$T& CC3M, CC12M& Auto. & \XSolidBrush &134M/--/--& 1 & 5K \\
StableLLaVA~\cite{li2023stablellava}& SFT & I+T$\to$T& SD~\cite{rombach2022high}& Auto.+Manu. & \XSolidBrush &126K/--/--& 1 & 126K \\

LLaVA's IT~\cite{zhang2023llavar}& SFT& I+T$\to$T&MS-COCO& Auto.&\CheckmarkBold&81K/--/--&2.29&150K \\

SVIT~\cite{zhao2023svit} & SFT& I+T$\to$T & MS-COCO, Visual Genome & Auto. & \CheckmarkBold & 108K/--/--& 5&3.2M\\

LLaVAR's IT~\cite{zhang2023llavar} & SFT &  I+T$\to$T & MS-COCO, CC3M, LAION & LLaVA+Auto. & \CheckmarkBold & 20K/--/-- & 2.27&174K\\

ShareGPT4V's IT~\cite{chen2023sharegpt4v} & SFT & I+T$\to$T & LCS, COCO, SAM, TextCaps, WikiArt & Auto.+Manu. & \XSolidBrush & 100K/--/-- & -- & -- \\

DRESS's IT~\cite{chen2023dress} & SFT& I+T$\to$T & LLaVA's IT, VLSafe & Auto.+Manu.& \CheckmarkBold & 193K/--/-- & $\sim$4 &-- \\

SoM-LLaVA's IT~\cite{yan2024list} & SFT& I+T$\to$T & ShareGPT4V's IT, LLaVA-1.5's IT, CogVLM's IT & Auto.+Manu.& \CheckmarkBold & --/--/-- & $\sim$5 &695K\\

VideoChat's IT~\cite{li2023videochat} & SFT & V+T$\to$T& WebVid& Auto. & \CheckmarkBold & --/8K/--& 1.82 & 11K\\

Video-ChatGPT's IT~\cite{maaz2023video} & SFT & V+T$\to$T& ActivityNet~\cite{caba2015activitynet} & Inherit & \CheckmarkBold & --/100K/-- & 1 & 100K \\

Video-LLaMA's IT~\cite{DBLP:conf/emnlp/ZhangLB23} & SFT & I/V+T$\to$T& MiniGPT-4, LLaVA, and VideoChat's IT & Auto. & \CheckmarkBold & 81K/8K/-- & 2.22 & 171K \\

InstructBLIP's IT~\cite{DBLP:journals/corr/abs-2305-06500} & SFT & I/V+T$\to$T & Multiple (InstructBLIP's Figure 2) & Auto. & \XSolidBrush & -- & -- & $\sim$1.6M \\

X-InstructBLIP's IT~\cite{panagopoulou2023x} & SFT & I/V/A/3D+T$\to$T & Multiple (X-InstructBLIP's Figure 4) & Auto. & \XSolidBrush & -- & -- & $\sim$1.8M \\

MIMIC-IT~\cite{li2023mimic} & SFT & I/V+T$\to$T & Multiple & Auto. & \XSolidBrush & 8.1M/502K/-- & 1 & 2.8M \\

PandaGPT's IT~\cite{su2023pandagpt} & SFT & I+T$\to$T & MiniGPT-4 and LLaVA's IT & Inherit & \CheckmarkBold & 81K/--/-- & 2.29 & 160K \\

MGVLID~\cite{zhao2023chatspot} & SFT & I+B+T$\to$T & Multiple & Auto.+Manu. & \XSolidBrush & 108K/--/-- & -- & 108K \\

M$^{\text{3}}$IT~\cite{li2023m} & SFT & I/V/B+T$\to$T & Multiple & Auto.+Manu. & \XSolidBrush & --/--/-- & 1 & 2.4M \\

LAMM~\cite{yin2023lamm} & SFT & I+3D+T$\to$T & Multiple & Auto.+Manu. & \CheckmarkBold & 91K/--/-- & 3.27 & 196K \\

BuboGPT's IT~\cite{zhao2023bubogpt} & SFT & (I+A)/A+T$\to$T & Clotho, VGGSS & Auto. & \XSolidBrush & 5K/--/9K & -- & 9K \\

mPLUG-DocOwl's IT~\cite{ye2023mplug} & SFT & I/Tab/Web+T$\to$T & Multiple & Inherit & \XSolidBrush & -- & -- & --\\

T2M~\cite{wu2023next} & SFT & T$\to$I/V/A+T & WebVid, CC3M, AudioCap & Auto. & \XSolidBrush & 4.9K/4.9K/4.9K & 1 & 14.7K\\

MosIT~\cite{wu2023next} & SFT & I+V+A+T$\to$I+V+A+T & Youtube, Google, Flickr30k, Midjourney, etc. & Auto.+Manu. & \CheckmarkBold & 4K/4K/4K & 4.8 & 5K  \\

Osprey's IT~\cite{yuan2023osprey} & SFT & I+T$\to$T & MS-COCO, RefCOCO, RefCOCO+, LLaVA's IT etc. (fine-grained region-text dataset) & Auto.+Manu. & \CheckmarkBold & --/--/-- & $\sim$4 &  724K  \\

LLaVA-RLHF~\cite{sun2023aligning} & RLHF& I+T$\to$T & Collected human preference  & Manu.& \XSolidBrush & --/--/-- & -- & 10K \\

DRESS's IT~\cite{chen2023dress} & RLHF& I+T$\to$T & LLaVA's IT, VLSafe & Auto.+Manu.& \CheckmarkBold & 33K/--/-- & $\sim$4 & -- \\

RLHF-V's IT~\cite{yu2023rlhf} & RLHF& I+T$\to$T & Collected human preference  & Manu.& \XSolidBrush & --/--/-- & -- & 1.4K \\

VLFeedback~\cite{li2023silkie} & RLHF & I+T$\to$T & Responses generated by $12$ MM-LLMs & Auto. & \XSolidBrush & --/--/-- & -- & 80K \\

RTVLM~\cite{li2024red} & RLHF & I+T$\to$T & \makecell{New question-image pairs based on publicly available images\\ or originally diffusion-generated images~\cite{gallegos2023bias}} & Auto.+Manu. & \XSolidBrush & --/--/-- & -- & 5K \\

VLGuard's IT~\cite{zong2024safety} & RLHF & I+T$\to$T & Source image data from various datasets & Auto. & \XSolidBrush & 3K/--/-- & -- & 3K \\

MMViG~\cite{yan2024vigor} & RLHF & I+T$\to$T & MS-COCO & Manu. & \XSolidBrush & 16K/--/-- & -- & 16K \\
 
\bottomrule
\end{tabular}
}
\caption{The statistics for MM IT datasets. I$\to$O: Input to Output Modalities, T: Text, I: Image, V: Video, A: Audio, B: Bounding box, 3D: Point Cloud, Tab: Table, and Web: Web page.}
\label{IT_corpus}
\end{table*}

\section{VL Benchmarks}
\label{ap:benchmark}

The $18$ VL benchmarks presented in Table~\ref{tab:benchmarks} include \textbf{OKVQA}~\cite{schwenk2022okvqa}, 
\textbf{IconVQA}~\cite{lu2021iconqa}, 
\textbf{VQA$^\text{v2}$}~\cite{goyal2017making}, 
\textbf{GQA}~\cite{hudson2019gqa}, 
\textbf{VizWiz}~\cite{gurari2018vizwiz}, 
\textbf{SQA$^\text{I}$}: ScienceQA-IMG~\cite{lu2022learn}, 
\textbf{VQA$^\text{T}$}: TextVQA~\cite{singh2019towards}, \textbf{POPE}~\cite{li2023evaluating}, 
\textbf{MME$^\text{P}$}: MME Perception~\cite{fu2023mme}, 
\textbf{MME$^\text{C}$}: MME Cognition~\cite{fu2023mme}, 
\textbf{MMB}: MMBenchmark~\cite{liu2023mmbench}, 
\textbf{MMB$^\text{CN}$}: MMBench-Chinese~\cite{liu2023mmbench}, \textbf{SEED$^\text{I}$}: SEED-Bench (Image)~\cite{li2023seed}, \textbf{LLaVA$^\text{W}$}: LLaVA-Bench (In-the-Wild)~\cite{liu2023visual}, 
\textbf{MM-Vet}~\cite{yu2023mm}, 
\textbf{QBench}~\cite{wu2023q}, 
\textbf{HM}: HatefulMemes~\cite{kiela2020hateful}, 
and \textbf{VSR}~\cite{liu2023visual}.

\section{Training Dataset}
\label{ap:corpus}

The statistics for MM PT and MM IT dataset are presented in Table~\ref{PT_corpus} and Table~\ref{IT_corpus}, respectively.

\end{document}